\documentclass[10pt,twocolumn,letterpaper]{article}

\usepackage{cvpr}              

\definecolor{cvprblue}{rgb}{0.21,0.49,0.74}
\usepackage[pagebackref,breaklinks,colorlinks,allcolors=cvprblue]{hyperref}

\usepackage{multirow}
\usepackage{colortbl}
\usepackage{diagbox}
\usepackage{tabularx}
\usepackage{stfloats}
\usepackage{graphicx}
\usepackage{capt-of}

\definecolor{jqcolor}{HTML}{1D953F}

\usepackage{xcolor}

\def\paperID{19061} 
\def\confName{CVPR}
\def\confYear{2026}

\title{Bootstrap Dynamic-Aware 3D Visual Representation for\\ Scalable Robot Learning}

\author{
Qiwei Liang\textsuperscript{1,2,$\ast$}, 
Boyang Cai\textsuperscript{1,2,$\ast$}, 
Minghao Lai\textsuperscript{2,$\ast$}, 
Sitong Zhuang\textsuperscript{2,$\ast$},
Tao Lin\textsuperscript{3}\\
Yan Qin\textsuperscript{1,2}, 
Yixuan Ye\textsuperscript{4}, 
Jiaming Liang\textsuperscript{1,2}, 
Renjing Xu\textsuperscript{1,$\dagger$} \\
\normalsize
\\[-0.35cm]
\textsuperscript{1}Hong Kong University of Science and Technology (Guangzhou) \\
\textsuperscript{2}Shenzhen University \quad
\textsuperscript{3}Beijing Jiaotong University \quad
\textsuperscript{4}Central South University \\
\normalsize
$^\ast$Equal contribution \quad $^\dagger$Corresponding author \\
}

\begin{document}

\makeatletter
\let\@oldmaketitle\@maketitle
\renewcommand{\@maketitle}{\@oldmaketitle
 \centering
    \includegraphics[width=\textwidth]{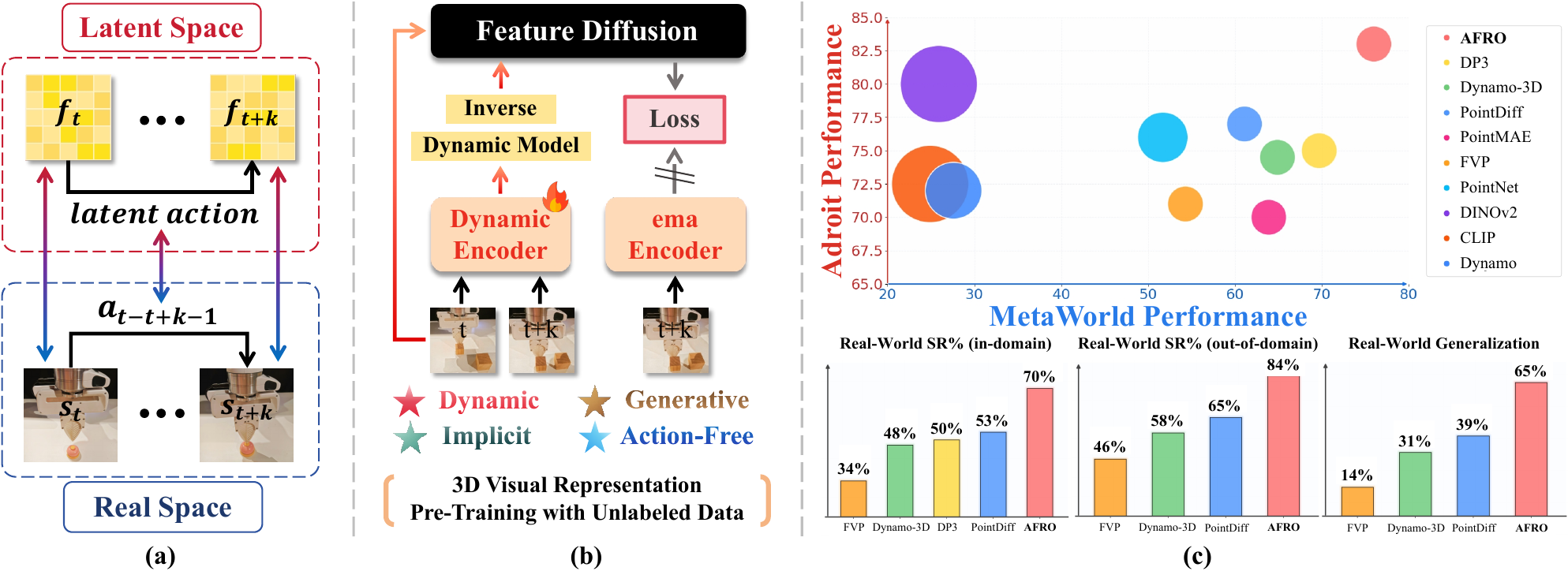}
    \vspace{-0.4cm}
    \captionof{figure}[]{(a) The relationship between robot manipulation in real space and its abstraction in latent space. (b) Our framework learns dynamics-aware 3D visual features in latent space, replacing static representations without relying on explicit action labels or reconstruction. (c) AFRO achieves higher success rates and stronger generalization than baseline methods in both simulation and real-world tasks.}
    \label{fig1:teaser}
  \bigskip}
\makeatother

\maketitle
\begin{abstract}
Despite strong results on recognition and segmentation, current 3D visual pre-training methods often underperform on robotic manipulation. We attribute this gap to two factors: the lack of state–action–state dynamics modeling and the unnecessary redundancy of explicit geometric reconstruction. We introduce \textbf{AFRO}, a self-supervised framework that learns dynamics-aware 3D representations without action or reconstruction supervision. \textbf{AFRO} casts state prediction as a generative diffusion process and jointly models forward and inverse dynamics in a shared latent space to capture causal transition structure. To prevent feature leakage in action learning, we employ feature differencing and inverse-consistency supervision, improving the quality and stability of visual features. When combined with Diffusion Policy, \textbf{AFRO} substantially increases manipulation success rates across 16 simulated and 4 real-world tasks, outperforming existing pre-training approaches. The framework also scales favorably with data volume and task complexity. Qualitative visualizations indicate that \textbf{AFRO} learns semantically rich, discriminative features, offering an effective pre-training solution for 3D representation learning in robotics. Project page: \url{https://kolakivy.github.io/AFRO/}
\end{abstract}

\section{Introduction}
Visual pre-training has significantly advanced robotic manipulation by decoupling representation learning from policy learning: models first acquire visual priors from large-scale datasets and then transfer them to downstream control tasks~\cite{RPR,hrp,sensorimotor,real-pre,unbiased-pre-training,exploring,learningact}. This paradigm enables richer features and cross-domain transfer. Recent 2D approaches (e.g., MVP~\cite{mvp_ref}, R3M~\cite{r3m_ref}, VIP~\cite{vip_ref}) leverage large-scale image or video corpora and substantially improve manipulation performance.

However, although 3D point cloud inputs provide richer geometric information and stronger generalization~\cite{rvt,whole,points}, 3D visual pre-training for robotic manipulation remains underexplored, even as standardized pipelines and generative tools rapidly expand the scale and quality of 3D data~\cite{robomind,rh20t}. These trends motivate a central question: \textbf{\emph{Can we design a scalable 3D visual pre-training framework tailored to robotic manipulation?}}

Point clouds provide a natural representation of 3D scenes, yet their irregular structure, sparsity, and permutation invariance complicate spatial reasoning and semantic abstraction~\cite{pointnet_ref}. Although many 3D pre-training methods perform well on object recognition~\cite{pointbert,strl} and segmentation~\cite{pointcontrast}, they transfer poorly to robotic manipulation and can even lag behind 2D foundation models on manipulation benchmarks~\cite{lift3d}. We identify two key limitations in existing 3D pre-training approaches:

\textbf{Lack of dynamics awareness.} Robotic manipulation is inherently sequential, where each action changes the system state. However, most 3D pre-training frameworks rely on supervision from single frames, ignoring temporal continuity and causal dependencies between states. As a result, the learned representations lack coherent temporal structure and fail to capture dynamic relationships across states.

\textbf{Lack of manipulation-relevant abstraction.} Many 3D methods focus on holistic scene reconstruction and often capture background details irrelevant to control~\cite{univla}. Such dense representations may mislead policy networks by diverting attention from task-critical elements. In contrast, effective manipulation requires abstractions that emphasize actionable object regions and interaction dynamics.

Robotic manipulation forms state–action–state trajectories rich in supervisory signals. Yet scalable visual representation learning often omits these labels, raising a key question: how can models capture dynamics without explicit action or transition annotations? We argue that dynamics can emerge through latent-space objectives. Recent latent-predictive learning approaches (e.g., V-JEPA 2~\cite{v-jepa2}) show that aligning high-level features without pixel-level reconstruction can produce generalizable representations.

Building on this, we introduce AFRO, a self-supervised 3D pre-training framework for learning dynamics-aware representations from point clouds in a world-modeling sense~\cite{ye2026mind}. AFRO aims (i) to model spatiotemporal relationships across states and (ii) to learn manipulation-relevant abstractions without explicit reconstruction.

Concretely, AFRO integrates inverse and forward dynamics models (IDM/FDM) to encode state–action–state transitions in latent space. As shown in Figure~\ref{fig1:teaser}, consecutive point clouds are encoded into features; the IDM infers a latent action from adjacent states, while the FDM predicts future features conditioned on the current feature and inferred action. State prediction occurs entirely in feature space, where the encoder is trained using a VICReg~\cite{vicreg} loss against an exponential moving average (EMA) target encoder. To capture multimodal futures, we model feature prediction as a diffusion-based generation process.

To prevent latent action shortcutting and feature leakage in IDM learning, we input feature differences instead of raw feature pairs, forcing the model to reason about action-driven change rather than memorizing states. We further introduce inverse-consistency supervision, applying the same pipeline in reverse to explain previous states from future observations. By requiring latent actions to support both forward prediction and backward explanation, we avoid degenerate solutions and stabilize representation learning.

We evaluate AFRO on 16 simulated and 4 real-world manipulation tasks spanning skills (e.g., sliding, pressing, pushing). Across benchmarks, AFRO outperforms 2D/3D pretraining baselines and imitation-from-scratch, with gains that scale with both data volume and domain diversity. Pretraining on the large-scale RH20T~\cite{rh20t} dataset further improves downstream results across all tasks.

Our core contributions are: \textbf{(1)} We propose a 3D visual pretraining framework for robotic manipulation that learns dynamics-aware representations directly in latent space, modeling future state uncertainty with diffusion while avoiding explicit reconstruction. \textbf{(2)} To our knowledge, we are the first to introduce latent actions into 3D visual learning and design feature-differencing and inverse-consistency supervision to prevent shortcut learning and feature leakage, improving representation quality and stability. \textbf{(3)} Extensive simulation and real-robot experiments demonstrate clear improvements over strong baselines, with visualizations and ablations validating each component.

\section{Related Work}
\label{sec:related work}
\subsection{Visual Pretraining for Robot Manipulation}

Visual pretraining significantly enhances data efficiency in robotic manipulation by learning transferable representations from large-scale visual data~\cite{embodiedmae,ar_vrm}. Early 2D approaches such as R3M~\cite{r3m_ref}, MVP~\cite{mvp_ref}, and VIP~\cite{vip_ref} leverage contrastive or masked modeling techniques to extract semantic pixel-level features. However, these methods are largely confined to image-space reasoning and lack explicit 3D geometric awareness, which is crucial for accurate spatial perception in complex real-world environments.

Recent work has shifted toward 3D visual pretraining to better capture scene geometry. For instance, 3D-MVP~\cite{3d-mvp} extends MVP by incorporating multi-view masked autoencoding on large-scale 3D data; SUGAR~\cite{sugar} jointly learns semantic and affordance representations from point clouds, improving zero-shot manipulation capabilities. However, most existing 3D methods still overlook the inherent temporal dynamics of robotic manipulation. Although FVP~\cite{fvp_ref} advances this direction with diffusion-based future prediction, it remains dependent on explicit scene reconstruction. In contrast, our approach AFRO introduces action-free 3D pretraining that learns dynamics-aware representations directly in latent space, bridging 3D perception and embodied behavior without relying on action labels or reconstruction.

\begin{figure*}[!t]
    \includegraphics[width=\textwidth]{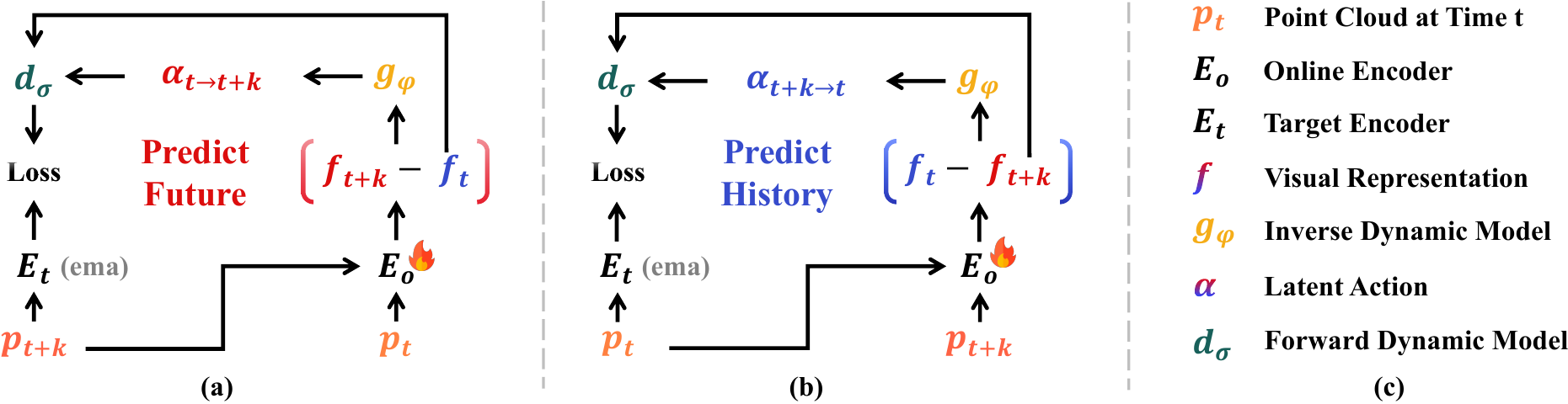}
    \centering
    \caption{Overall framework of our method.
    \textbf{(a)} Predict Future: Given point clouds $\mathcal{P}_{t}$ and $\mathcal{P}_{t+k}$, the online encoder $f_{\phi}$ produces features $\mathbf{z}_{t}$ and $\mathbf{z}_{t+k}$. The inverse dynamics model $g_{\psi}$ takes the difference $(\mathbf{z}_{t+k}-\mathbf{z}_{t})$ to infer the latent action $\boldsymbol{\alpha}_{t\rightarrow t+k}$. The target encoder $f_{\xi}$ (EMA) encodes $\mathcal{P}_{t+k}$ to obtain $\tilde{\mathbf{z}}_{t+k}$. The forward dynamics model $h_{\theta}$ predicts $(\mathbf{z}_{t}, \boldsymbol{\alpha}_{t\rightarrow t+k}) \mapsto \hat{\mathbf{z}}_{t+k}$ and aligns it with $\tilde{\mathbf{z}}_{t+k}$, guiding $f_{\phi}$ to learn dynamics-aware representations.
    \textbf{(b)} Predict History: Using $(\mathbf{z}_{t}-\mathbf{z}_{t+k})$, $g_{\psi}$ infers $\boldsymbol{\alpha}_{t+k\rightarrow t}$. The encoder $f_{\xi}$ encodes $\mathcal{P}_{t}$ to obtain $\tilde{\mathbf{z}}_{t}$, while $h_{\theta}$ predicts $(\mathbf{z}_{t+k}, \boldsymbol{\alpha}_{t+k\rightarrow t}) \mapsto \hat{\mathbf{z}}_{t}$ aligned with $\tilde{\mathbf{z}}_{t}$. Other steps are symmetric to (a).
    \textbf{(c)} Notation overview.}
    \label{fig1 teaser}
\end{figure*}

\subsection{Latent Actions for Robot Learning}

Latent actions encode state transitions into low-dimensional variables, offering an alternative to expensive action annotations~\cite{latent-supervision, taco, latent-behavior, adaworld, allshire2021laser}. Early work employed inverse dynamics models (IDM) for pretraining, while later approaches jointly learned IDM and forward dynamics models (FDM) in visual spaces to improve sample efficiency~\cite{villa, AMPLIFY, latent-world, jif}. Current research falls into three directions: \textbf{(1)} using latent actions as policy supervision from unlabeled videos, e.g., CLAM~\cite{clam} and CoMo~\cite{como}; \textbf{(2)} vision-language-action pretraining, where LAPA~\cite{lapa} discretizes frame changes into tokens and UniVLA~\cite{univla} learns task-centric latent actions in DINO space for cross-embodiment generalization; and \textbf{(3)} reinforcement learning with constrained action spaces to address OOD issues, PLAS~\cite{plas}.

Recently, Dynamo~\cite{dynamo_ref} incorporates latent actions into visual pretraining to explicitly capture dynamic changes in manipulation. However, existing approaches remain largely limited to 2D visual spaces. In contrast, we introduce latent actions into 3D pretraining and mitigate IDM’s tendency to overfit future features by (i) using feature differencing instead of frame stacking as IDM input, and (ii) introducing inverse-consistency supervision that enables FDM to predict historical states from future observations, improving reversibility and better temporal consistency.

\section{Preliminary}
\label{sec:preliminary}

We aim to learn dynamics-aware 3D representations from unlabeled point cloud demonstrations. Our approach adopts a latent dynamics paradigm with an online encoder $f_{\phi}$, inverse dynamics model (IDM) $g_{\psi}$, and forward dynamics model (FDM) $h_{\theta}$. Given consecutive point clouds $\mathcal{P}_{t}$ and $\mathcal{P}_{t+k}$, the encoder extracts features $\mathbf{z}_{t}$ and $\mathbf{z}_{t+k}$. The IDM infers a latent action
\begin{equation}
\boldsymbol{\alpha}_{t \rightarrow t+k} = g_{\psi}\big(\mathbf{z}_{t}, \mathbf{z}_{t+k}\big),
\end{equation}
and the FDM predicts the future feature
\begin{equation}
\hat{\mathbf{z}}_{t+k} = h_{\theta}\big(\mathbf{z}_{t}, \boldsymbol{\alpha}_{t \rightarrow t+k}\big),
\end{equation}
enabling state prediction in latent space, while an EMA-updated encoder $f_{\xi}$ provides stable targets $\tilde{\mathbf{z}}$.

However, two limitations remain. First, the IDM may suffer from feature leakage, where shortcuts encode information from $\mathbf{z}_{t+k}$ and produce degenerate latent actions. Second, standard FDMs are deterministic and struggle with multimodal uncertainty in real-world interactions, often predicting averaged futures. To address these issues, we introduce two components in Sec.~\ref{sec:method}:
\begin{enumerate}
\item \textbf{feature-differencing and reverse-consistency} framework for latent action learning that prevents feature leakage and improves temporal coherence (Sec.~\ref{subsec:latent_action}).
\item \textbf{diffusion-based forward dynamics model} that captures multimodal future distributions (Sec.~\ref{subsec:diffusion_fdm}).
\end{enumerate}

\section{Method}
\label{sec:method}
\textbf{Overview.} As shown in Fig.~\ref{fig1 teaser}, our framework learns 3D dynamics-aware representations from unlabeled point cloud sequences without reconstruction. It includes three parts: Sec.~\ref{subsec:latent_action} introduces latent action modeling, Sec.~\ref{subsec:diffusion_fdm} presents the diffusion-based forward dynamics, and Sec.~\ref{subsec:vicreg} describes the representation matching objective.

\begin{figure}[t]
    \centering
    \includegraphics[width=0.95\linewidth]{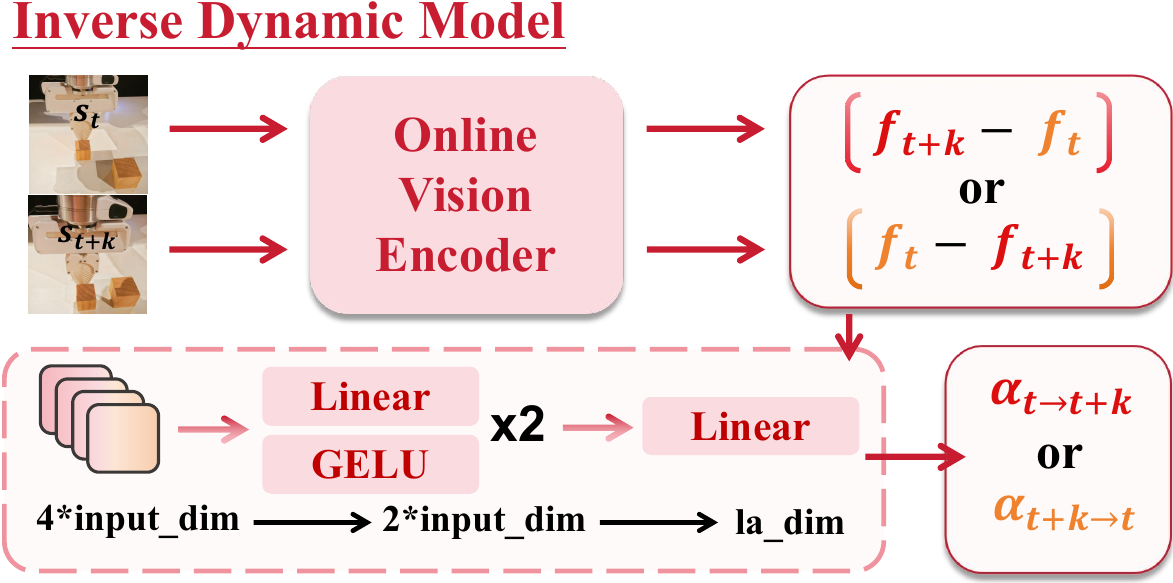}
    \caption{Given differential features \((\mathbf{z}_{t+k}-\mathbf{z}_{t})\) or \((\mathbf{z}_{t}-\mathbf{z}_{t+k})\), the inverse dynamics model \(g_{\psi}\) applies two Linear–GELU layers and a final Linear projection to produce latent actions \(\boldsymbol{\alpha}_{t\rightarrow t+k}\) or \(\boldsymbol{\alpha}_{t+k\rightarrow t}\), encoding motion across states.}
    \label{fig3:IDM}
\end{figure}

\subsection{Latent Action Modeling}
\label{subsec:latent_action}

Naively training inverse and forward dynamics on consecutive features \(\mathbf{z}_{t}\) and \(\mathbf{z}_{t+k}\) often causes feature leakage. Without explicit action supervision, the inverse model can reduce its loss by implicitly copying information from \(\mathbf{z}_{t+k}\) instead of reasoning about how the system transitioned between states. This shortcut undermines temporal reasoning, prevents the model from discovering meaningful dynamics, and ultimately produces degenerate latent actions.

We model changes in representations rather than the representations themselves. Concretely, instead of feeding \(\mathbf{z}_{t}\) and \(\mathbf{z}_{t+k}\) into the inverse model, we take their difference and learn latent actions directly from the displacement:
\begin{equation}
\boldsymbol{\alpha}_{t \rightarrow t+k} = g_{\psi}\!\big(\mathbf{z}_{t+k} - \mathbf{z}_{t}\big),
\end{equation}
where \(g_{\psi}\) is a lightweight MLP that maps the feature difference into a compact latent action space. This differential view emphasizes what changed and de-emphasizes what stayed the same, naturally filtering out static scene content and highlighting motion-relevant cues. In practice, we find this simple choice stabilizes training and makes the learned actions easier to interpret.

Beyond differencing, we also introduce an \textbf{inverse-consistency supervision} to promote temporal coherence. We define the inverse mapping symmetrically as
\begin{equation}
\boldsymbol{\alpha}_{t+k \rightarrow t} = g_{\psi}\!\big(\mathbf{z}_{t} - \mathbf{z}_{t+k}\big),
\end{equation}
encouraging the model to learn transformations that are approximately reversible and physically plausible. The forward dynamics model \(h_{\theta}\) is then asked to reconstruct the current representation from the future one:
\begin{equation}
\hat{\mathbf{z}}_{t} = h_{\theta}\!\big(\mathbf{z}_{t+k},\, \boldsymbol{\alpha}_{t+k \rightarrow t}\big),
\end{equation}
which aligns the two directions and constrains latent actions to be causally consistent. Differencing plus inverse-consistency provides a lightweight effective recipe for learning latent actions without explicit labels.

\begin{figure}[t]
    \centering
    \includegraphics[width=0.95\linewidth]{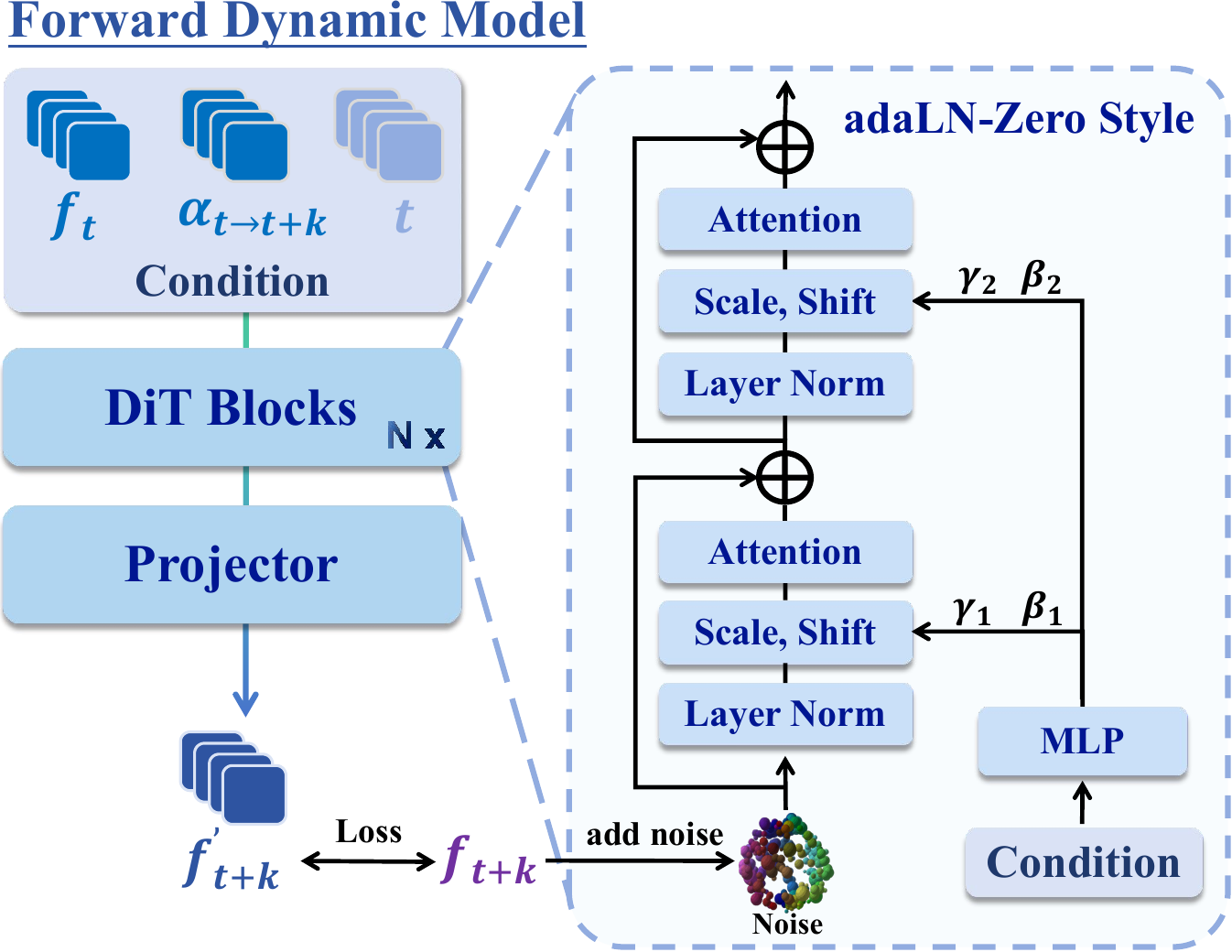}
    \caption{Architecture of the Forward Dynamic Model. An AdaLN-Zero diffusion transformer \(h_{\theta}\) conditions on \(\mathbf{z}_{t}\), latent action \(\boldsymbol{\alpha}_{t\rightarrow t+k}\), and timestep \(\tau\) encoded via MLPs and concatenation, which modulate LayerNorm through adaptive scale–shift. Attention and projector modules then denoise the latent to reconstruct the future representation \(\hat{\mathbf{z}}_{t+k}\).}
    \label{fig4:FDM}
\end{figure}

\subsection{Forward Dynamics with Diffusion Transformer}
\label{subsec:diffusion_fdm}

The future is uncertain and multimodal, as occlusions and stochastic interactions produce multiple plausible outcomes. Conventional MLP- or transformer-based predictors struggle to capture such uncertainty and often average over modes. We instead formulate forward prediction as conditional denoising: given the current feature $\mathbf{z}_t$, latent action $\boldsymbol{\alpha}_{t\rightarrow t+k}$, and diffusion step $\tau$, the forward dynamics model $h_{\theta}$—implemented as a diffusion transformer~\cite{dit} (DiT) with AdaLN-Zero conditioning—maps a noisy sample to a clean future latent representation.

\textbf{Noising (forward process).} We corrupt the clean future feature with a standard diffusion forward process:
\begin{equation}
\mathbf{z}^{(\tau)}_{t+k}
= \sqrt{\bar{\alpha}_{\tau}}\;\mathbf{z}_{t+k}
+ \sqrt{1-\bar{\alpha}_{\tau}}\;\boldsymbol{\epsilon},
\quad \boldsymbol{\epsilon}\!\sim\!\mathcal{N}(\mathbf{0},\mathbf{I}),
\label{eq:fdm_noise_only}
\end{equation}
where \(\bar{\alpha}_{\tau}\!\in\!(0,1]\) is the cumulative signal weight (equivalently \(\mathrm{SNR}(\tau)=\bar{\alpha}_{\tau}/(1-\bar{\alpha}_{\tau})\)).

\textbf{Direct denoising (student).} The noisy latent is mapped to a clean target in a single pass:
\begin{equation}
\hat{\mathbf{z}}_{t+k}
= h_{\theta}\!\big(\mathbf{z}^{(\tau)}_{t+k},\;\mathbf{z}_t,\;\boldsymbol{\alpha}_{t\rightarrow t+k},\;\tau\big).
\label{eq:fdm_single_step_gen}
\end{equation}
The noisy input and condition \((\mathbf{z}_t,\boldsymbol{\alpha}_{t\rightarrow t+k},\tau)\) pass through DiT blocks and a projector to produce \(\hat{\mathbf{z}}_{t+k}\).

\textbf{AdaLN-Zero conditioning.} Each DiT block applies identity-preserving adaptive LayerNorm
\(\mathrm{AdaLN}(\mathbf{x};\mathbf{c})=\mathrm{LN}(\mathbf{x})\odot\big(1+\gamma(\mathbf{c})\big)+\beta(\mathbf{c})\),
with \(\mathbf{c}=\mathrm{cond}(\mathbf{z}_t,\boldsymbol{\alpha}_{t\rightarrow t+k},\tau)\).
The scale and shift, \(\gamma(\cdot)\) and \(\beta(\cdot)\), are produced by small MLPs and zero-initialized, keeping the residual path near identity while progressively injecting timestep and action information—enabling the one-shot denoising in Eq.~\eqref{eq:fdm_single_step_gen}.

\subsection{VICReg Matching}
\label{subsec:vicreg}
To prevent collapse in self-supervised latent learning, we adopt Variance–Invariance–Covariance Regularization (VICReg)~\cite{vicreg}. Given predicted features \(\hat{\mathbf{z}}\) and teacher targets \(\tilde{\mathbf{z}}\) from the EMA encoder \(f_{\xi}\), the loss combines three terms:
\begin{equation}
\mathcal{L}_{\text{VICReg}} =
\lambda_I \mathcal{L}_{\text{inv}} +
\lambda_V \mathcal{L}_{\text{var}} +
\lambda_C \mathcal{L}_{\text{cov}}.
\label{eq:vicreg}
\end{equation}
Here, \(\mathcal{L}_{\text{inv}}\) aligns student and teacher features, \(\mathcal{L}_{\text{var}}\) maintains variance, and \(\mathcal{L}_{\text{cov}}\) reduces cross-channel correlations, stabilizing latent dynamics and improving representations.

\subsection{Pre-training Details}
\label{subsec:pretraining}
We train AFRO on simulated and real 3D manipulation data. The inverse dynamics module $g_{\psi}$ is a 3-layer MLP with GeLU activation, outputting 16-dimensional latent actions. The forward dynamics model $h_{\theta}$ is a 4-layer Transformer with adaptive conditioning. We use frame interval $k=4$ and train for 300 epochs using AdamW (lr=$1\times10^{-4}$). The VICReg loss uses weights $\lambda_I{=}\lambda_V{=}25$, $\lambda_C{=}1$, and sets the variance threshold $\gamma{=}1$. The target encoder $f_{\xi}$ updates via EMA with a momentum starting at 0.996.

\begin{table*}[t]
\caption{
\textbf{Comparison on simulation benchmarks.} 
Type denotes the pretraining category:  
\textbf{2D-LT} = large-scale 2D pretraining;  
\textbf{3D-LT} = large-scale 3D pretraining;  
\textbf{3D-Pre-S} = static 3D pretraining without dynamic modeling;  
\textbf{3D-Pre-D} = dynamic-aware 3D pretraining with temporal modeling;   
Input Type: \textbf{RGB} = image input; \textbf{PC} = point cloud input.  
\textbf{Conclusion:} AFRO achieves the highest success rate across task categories, 
demonstrating our pre-training method can substantially enhance performance in robot manipulation.
}
\centering
\resizebox{0.99\textwidth}{!}{
\setlength{\tabcolsep}{4pt}
\renewcommand{\arraystretch}{0.9}
\begin{tabular}{lcccccccccc}
\toprule
\multirow{2}{*}{\textbf{Method}} &
\multirow{2}{*}{\textbf{Type}} &
\multirow{2}{*}{\textbf{Input Type}} &
\multicolumn{5}{c}{\textbf{MetaWorld}} &
\multicolumn{3}{c}{\textbf{Adroit}} \\
\cmidrule(lr){4-8} \cmidrule(lr){9-11}
 & & & Easy (3) & Medium (6) & Hard (2) & Very Hard (3) & \textbf{Mean S.R.} & Door & Pen & \textbf{Mean S.R.} \\
\midrule
CLIP~\cite{clip_ref}      & 2D-LT & RGB      & 29.3 & 14.0 & 5.0 & 55.3 & \multicolumn{1}{l}{24.9 \textbf{\textcolor[HTML]{C00000}{{($\downarrow$ 51.1)}}}} & 61.0 & 84.0 & \multicolumn{1}{l}{72.5 \textbf{\textcolor[HTML]{C00000}{{($\downarrow$ 10.5)}}}} \\
DINOv2~\cite{dinov2_ref}    & 2D-LT & RGB      & 34.0 & 14.0 & 5.0 & 55.3 &  \multicolumn{1}{l}{25.9 \textbf{\textcolor[HTML]{C00000}{{($\downarrow$ 50.1)}}}} & 76.0 & 84.0 & \multicolumn{1}{l}{80.0 \textbf{\textcolor[HTML]{C00000}{{($\downarrow$ 3.0)}}}} \\
\midrule
PointNet~\cite{pointnet_ref}  & 3D-LT & PC   & 71.3 & 49.7 & 23.0 & 55.3 &  \multicolumn{1}{l}{51.7 \textbf{\textcolor[HTML]{C00000}{{($\downarrow$ 24.3)}}}} & 80.0 & 72.0 & \multicolumn{1}{l}{76.0 \textbf{\textcolor[HTML]{C00000}{{($\downarrow$ 7.0)}}}} \\
\midrule
PointMAE~\cite{pointmae_ref}  & 3D-Pre-S & PC       & 77.3 & 61.3 & 42.0 & 70.0 &  \multicolumn{1}{l}{63.9 \textbf{\textcolor[HTML]{C00000}{{($\downarrow$ 12.1)}}}} & 64.0 & 76.0 & \multicolumn{1}{l}{70.0 \textbf{\textcolor[HTML]{C00000}{{($\downarrow$ 13.0)}}}} \\
PointDif~\cite{pointdiff_ref} & 3D-Pre-S & PC       & 83.3 & 58.0 & 41.0 & 58.7 &  \multicolumn{1}{l}{61.1 \textbf{\textcolor[HTML]{C00000}{{($\downarrow$ 14.9)}}}} & 76.0 & 78.0 & \multicolumn{1}{l}{77.0 \textbf{\textcolor[HTML]{C00000}{{($\downarrow$ 6.0)}}}} \\
\midrule
DynaMo~\cite{dynamo_ref}    & 2D-Pre-D & RGB     & 42.0 & 21.7 & 14.0 & 34.0 &  \multicolumn{1}{l}{27.6 \textbf{\textcolor[HTML]{C00000}{{($\downarrow$ 48.4)}}}} & 76.0 & 68.0 & \multicolumn{1}{l}{72.0 \textbf{\textcolor[HTML]{C00000}{{($\downarrow$ 11.0)}}}} \\
\midrule
DynaMo-3D~\cite{dynamo_ref} & 3D-Pre-D & PC       & 84.7 & 55.3 & 41.0 & 80.0 &  \multicolumn{1}{l}{64.9 \textbf{\textcolor[HTML]{C00000}{{($\downarrow$ 11.1)}}}} & 73.0 & 76.0 & \multicolumn{1}{l}{74.5 \textbf{\textcolor[HTML]{C00000}{{($\downarrow$ 8.5)}}}} \\
FVP~\cite{fvp_ref}       & 3D-Pre-D & PC       & 80.0 & 44.3 & 34.0 & 62.0 &  \multicolumn{1}{l}{54.3 \textbf{\textcolor[HTML]{C00000}{{($\downarrow$ 21.7)}}}} & 66.0 & 76.0 & \multicolumn{1}{l}{71.0 \textbf{\textcolor[HTML]{C00000}{{($\downarrow$ 12.0)}}}} \\
\midrule
DP3~\cite{dp3_ref}       & 3D-Pol & PC      & 82.7 & 65.0 & 49.0 & 80.0 &  \multicolumn{1}{l}{69.7 \textbf{\textcolor[HTML]{C00000}{{($\downarrow$ 6.3)}}}} & 70.0 & 80.0 & \multicolumn{1}{l}{75.0 \textbf{\textcolor[HTML]{C00000}{{($\downarrow$ 8.0)}}}} \\
\midrule
\rowcolor[HTML]{F9F9F9}
\textbf{{AFRO (Ours)}}& \textbf{3D-Pre-D} & \textbf{PC} & \textbf{88.0} & \textbf{69.7} & \textbf{55.0} & \textbf{90.7} & \textbf{76.0} & \textbf{82.0} & \textbf{84.0} & \textbf{83.0} \\
\bottomrule
\end{tabular}
}

\label{tab:afro-table}
\end{table*}

\section{Simulation Experiment}

\subsection{Simulation Benchmarks.}

\hspace*{1.35em}\textbf{Adroit}~\cite{adroit_ref} is a multi-DoF anthropomorphic hand benchmark in MuJoCo focused on fine-grained control. We evaluate two tasks, Door and Pen, both of which require precise manipulation and temporal coordination. 

\textbf{MetaWorld}~\cite{metaworld_ref} provides a diverse set of manipulation tasks using a 7-DoF Sawyer arm in MuJoCo. We select 14 representative tasks (all details in Appendix).

In MetaWorld, each task includes 25 expert trajectories generated using scripted policies, while in Adroit each task contains 100 expert trajectories produced by VRL3~\cite{vrl3_ref}. 

\noindent
\subsection{Baseline methods.}
To evaluate the effectiveness of AFRO, we compare it against four categories of baselines. (1) \textbf{End-to-End Learning} (DP3)~\cite{dp3_ref} jointly learns perception and control without visual pretraining. (2) \textbf{Large-Scale Dataset Pretraining} employs visual feature extractors such as CLIP~\cite{clip_ref}, DINOv2~\cite{dinov2_ref}, and PointNet~\cite{pointnet_ref} trained on large external datasets to improve generalization before diffusion policy training. (3) \textbf{Static-Only Pretraining} baselines, including Point-MAE~\cite{pointmae_ref} and PointDif~\cite{pointdiff_ref}, rely solely on static data, ignoring dynamic state relationships. (4) \textbf{Dynamic-Aware Pretraining} baselines, such as FVP~\cite{fvp_ref}, DynaMo-2D~\cite{dynamo_ref}, and DynaMo-3D, explicitly model temporal dynamics during visual pretraining, with DynaMo-3D being our 3D extension of the original DynaMo method.

\noindent
\subsection{Implementation Details.}
After visual pretraining, the encoder parameters were frozen for policy learning. We employed a diffusion policy architecture for all evaluations. Policies were trained for 100 epochs using the AdamW optimizer ($\beta_1, \beta_2 = 0.9, 0.999$), a learning rate of $1\mathrm{e}{-4}$ with cosine annealing, and a batch size of 128. Evaluation was conducted every 10 epochs with 25 rollouts, and we report the highest success rate among the top three performing policies. All baselines follow the same training and evaluation protocol, with experiments conducted on a single NVIDIA RTX 4090 GPU.

\subsection{Quantitative Results}
\label{subsec:quantitative}
As shown in Table~\ref{tab:afro-table}, \textsc{AFRO} achieves the best performance on both benchmarks under a unified protocol. 2D models (CLIP, DINOv2) provide semantic priors but perform poorly on MetaWorld (24.9\%, 25.9\%). Among 3D methods, static approaches (PointMAE, PointDif) benefit from geometric reconstruction but lack temporal reasoning, while dynamic baselines such as DynaMo-3D underperform static models. In contrast, \textsc{AFRO} combines diffusion-based future prediction with latent-action learning, reaching \textbf{76.0\%} success on MetaWorld (+6.3 over DP3) and \textbf{83.0\%} on Adroit. These gains suggest feature differencing and inverse-consistency supervision yield stable representations capturing causal state transitions.

\begin{figure*}[!t]
    \centering
    \includegraphics[width=\textwidth]{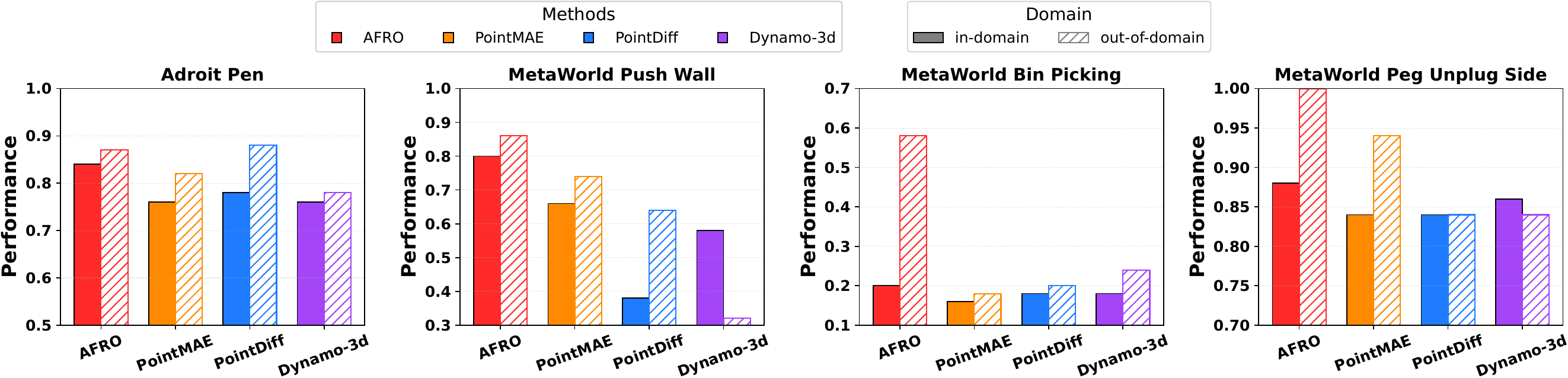}
    \caption{\textbf{Scaling across task domains.}
    Comparison of \emph{in-domain} (solid) and \emph{multi-domain} (hatched) visual pretraining.
    AFRO benefits from multi-domain data and reaches $100\%$ success on Peg Unplug Side, while baselines show smaller or inconsistent gains.}
    \label{fig:domain-scale}
\end{figure*}

\subsection{Scalability Analysis}

\begin{figure}[t]
    \centering
    \includegraphics[width=\linewidth]{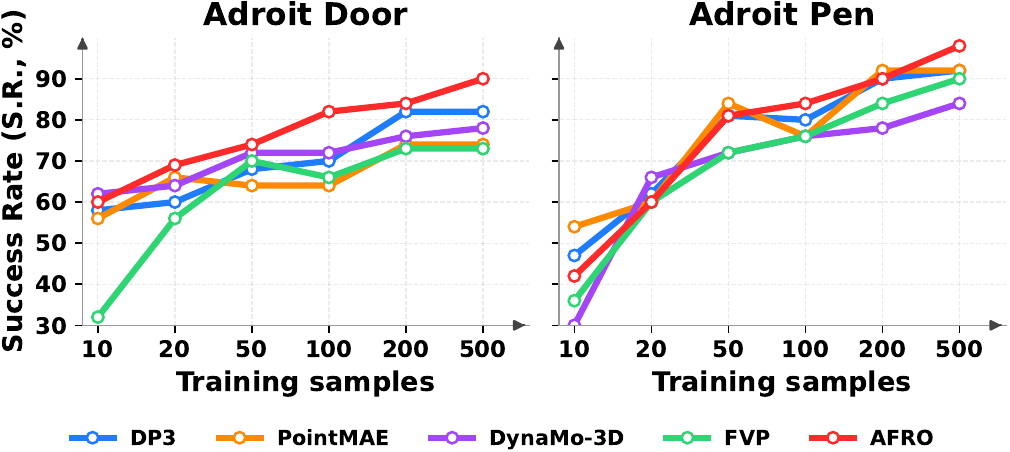}
    \caption{\textbf{Scaling with data.} Success rate versus number of expert trajectories ({10, 20, 50, 100, 200, 500}) for Door and Pen. 
    }
    \label{fig:data-scale}
\end{figure}

\noindent
\textbf{Scalable Evaluation of Domain.}
We study how pretraining scales with task diversity. The encoder is pretrained once on the union of tasks (Door+Pen for Adroit, 14 tasks for MetaWorld) and then frozen while a diffusion policy is trained for each task. As shown in Fig.~\ref{fig:domain-scale}, AFRO benefits from multi-domain pretraining: mean success rises from \textbf{68.0\%} to \textbf{82.8\%}, with Bin Picking improving from \textbf{20\%} to \textbf{58\%} and Peg Unplug Side reaching \textbf{100\%}. Other methods show smaller gains or degradation. This indicates AFRO learns transferable 3D dynamics by organizing states through transition structure instead of appearance. \textbf{Overall, AFRO scales favorably with increasing task diversity and heterogeneous training domains.}

\begin{figure}[t]
    \centering
    \includegraphics[width=\linewidth]{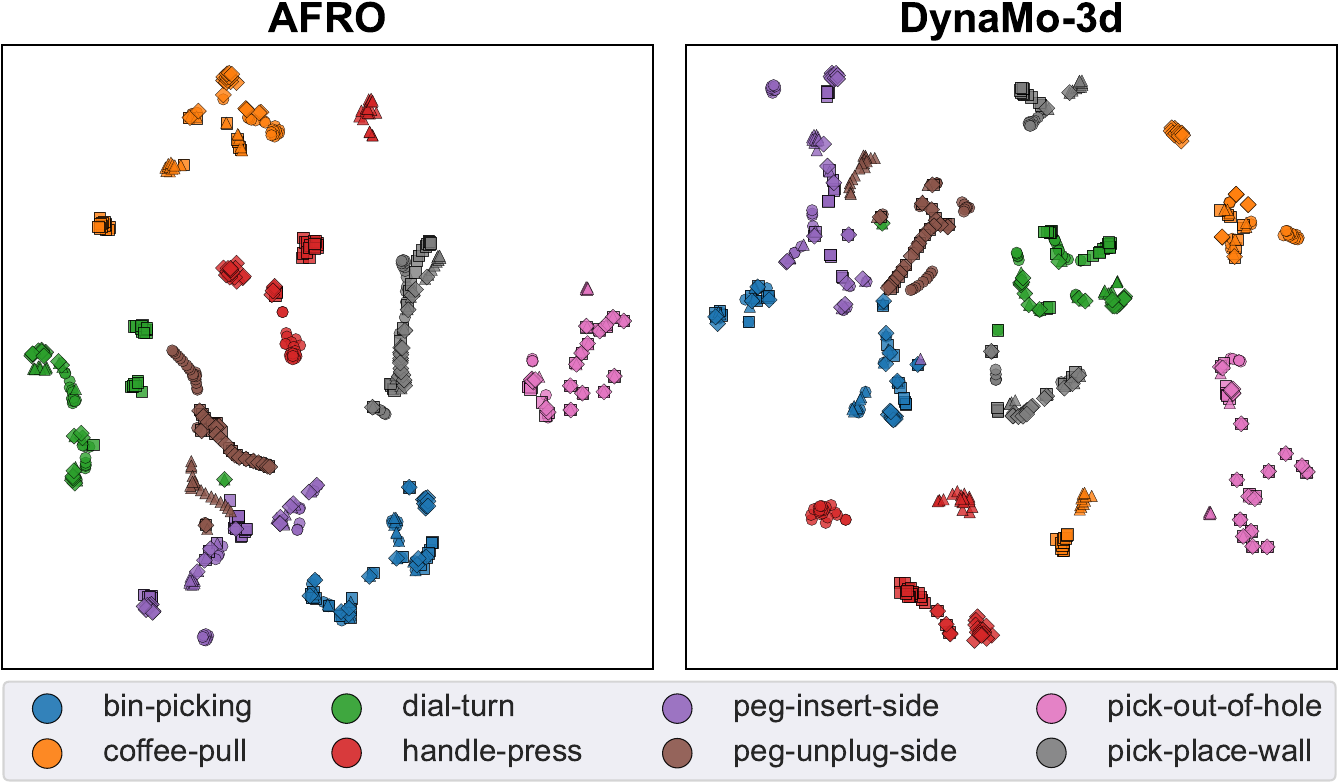}
    \caption{\textbf{t-SNE of latent features on eight MetaWorld tasks.}
    Colors denote tasks and marker shapes denote sequences.
    AFRO (left) forms clearer clusters than DynaMo-3D (right), indicating more discriminative and temporally coherent representations.}
    \label{fig:tsne}
\end{figure}

\noindent
\textbf{Scalable Evaluation of Data.}
We vary expert trajectories per task $\{10,20,50,100,200,500\}$ on Adroit-Door and Adroit-Pen (Fig.~\ref{fig:data-scale}). AFRO leads across all scales. On Door, it reaches \textbf{74\%} with 50 samples and \textbf{90\%} at 500, exceeding DP3 and static baselines that plateau around 70--76\%. On Pen, AFRO rises from 60\% to \textbf{81\%} when samples increase from 20 to 50, and to \textbf{98\%} at 500, while alternatives fluctuate or saturate early. This suggests AFRO leverages additional data to learn richer latent dynamics for more effective policy decoding. \textbf{Overall, AFRO exhibits consistently favorable scaling behavior as data increases.} 

\subsection{Representation Analysis} We visualize encoder features with t-SNE on eight MetaWorld tasks after out-of-domain pre-training (Fig.~\ref{fig:tsne}). DynaMo-3D features are more entangled across tasks and exhibit fragmented trajectories, indicating weaker dynamic organization in latent space. In contrast, AFRO yields well-separated task clusters and smooth trajectories within each task (different marker shapes), showing both strong task discrimination and coherent temporal evolution.

\begin{table*}[t]
\caption{\textbf{In-domain vs.\ out-of-domain real-world performance.}
Both blocks show success rates on the same four manipulation tasks.
The \textbf{\textcolor[HTML]{0070C0}{blue}} block uses in-domain pre-training, and the \textbf{\textcolor[HTML]{C00000}{red}} block uses pre-training on a large out-of-domain dataset before task-specific fine-tuning.
Orange numbers indicate the mean-success drop relative to AFRO.}
\centering
\setlength{\tabcolsep}{10pt}
\renewcommand{\arraystretch}{0.9}
\begin{tabular}{lccccc@{\hspace{14pt}}ccccc}
\toprule
 & \multicolumn{5}{c}{\textbf{\textcolor[HTML]{0070C0}{In-domain (Real-World)}}}
 & \multicolumn{5}{c}{\textbf{\textcolor[HTML]{C00000}{Out-of-domain (Pre-train \& FT)}}} \\
\cmidrule(lr){2-6}\cmidrule(lr){7-11}
\textbf{Method}
 & FPP & BP & CB & B2BA & \textbf{Mean S.R.}
 & FPP & BP & CB & B2BA & \textbf{Mean S.R.} \\
\midrule
DP3
 & 45 & 60 & 45 & 50 & 50\,\textcolor[HTML]{ED7D31}{($\downarrow 20$)}
 & -- & -- & -- & -- & -- \\
PointDif
 & 40 & 55 & 60 & 55 & 53\,\textcolor[HTML]{ED7D31}{($\downarrow 17$)}
 & 55 & 75 & 70 & 60 & 65\,\textcolor[HTML]{ED7D31}{($\downarrow 19$)} \\
FVP
 & 25 & 40 & 40 & 30 & 34\,\textcolor[HTML]{ED7D31}{($\downarrow 36$)}
 & 40 & 55 & 50 & 40 & 46\,\textcolor[HTML]{ED7D31}{($\downarrow 38$)} \\
DynaMo-3D
 & 45 & 40 & 55 & 50 & 48\,\textcolor[HTML]{ED7D31}{($\downarrow 22$)}
 & 50 & 50 & 65 & 65 & 58\,\textcolor[HTML]{ED7D31}{($\downarrow 26$)} \\
\rowcolor[HTML]{F9F9F9}
\textbf{AFRO (Ours)}
 & \textbf{65} & \textbf{75} & \textbf{70} & \textbf{70} & \textbf{70}
 & \textbf{75} & \textbf{90} & \textbf{85} & \textbf{85} & \textbf{84} \\
\bottomrule
\end{tabular}
\label{tab:real-world-in-out}
\end{table*}

\section{Real-World Experiment}

\begin{figure}[t]
    \centering
    \includegraphics[width=\linewidth]{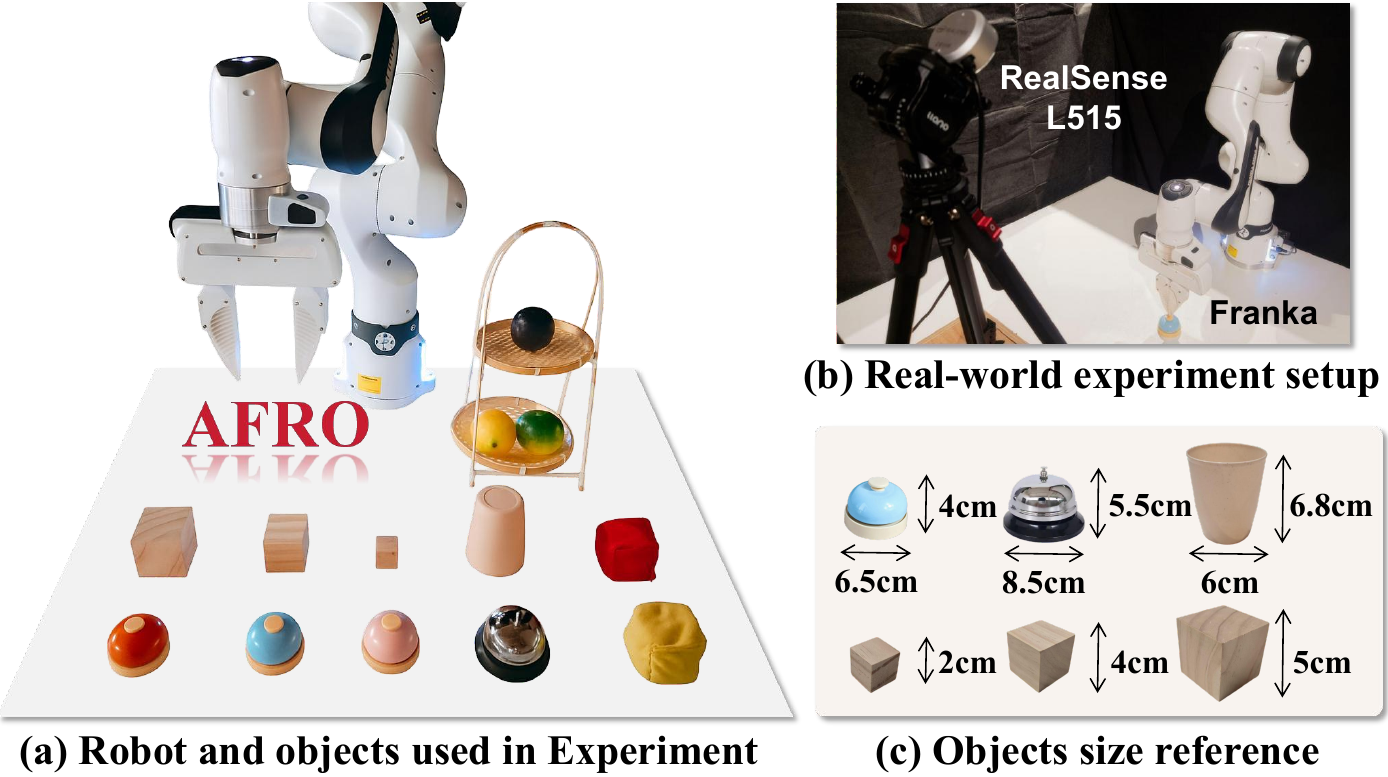}
    \caption{\textbf{Real-world evaluation setup and objects.} }
    \label{fig:real-word setup}
\end{figure}

\label{sec:real-world experiment}
\subsection{Experimental Setup}
We evaluate on a platform with a 7-DoF \textbf{Franka Emika} arm, parallel gripper, and a top-down \textbf{RealSense L515} depth camera; object layouts and camera/robot placement follow Fig.~\ref{fig:real-word setup}. We consider four manipulation tasks covering non-prehensile and prehensile control, perception precision, and kinematic difficulty:
\begin{itemize}
    \item \textbf{Block-to-Block Alignment}: The robot pushes a movable block to align it with a reference block. \emph{(B2BA)}
    
    \item \textbf{Bell Pressing}: The robot localizes a bell and presses it to trigger a click. \emph{(BP)}
    
    \item \textbf{Fruit Pick-and-Place}: The robot grasps a fruit from a random pose and places it into a target basket. \emph{(FPP)}
    
    \item \textbf{Cover Block}: The robot picks up a cup, moves above a target block, and places the cup to cover it. \emph{(CB)}
\end{itemize}

\begin{figure}[t]
    \centering
    \includegraphics[width=\linewidth]{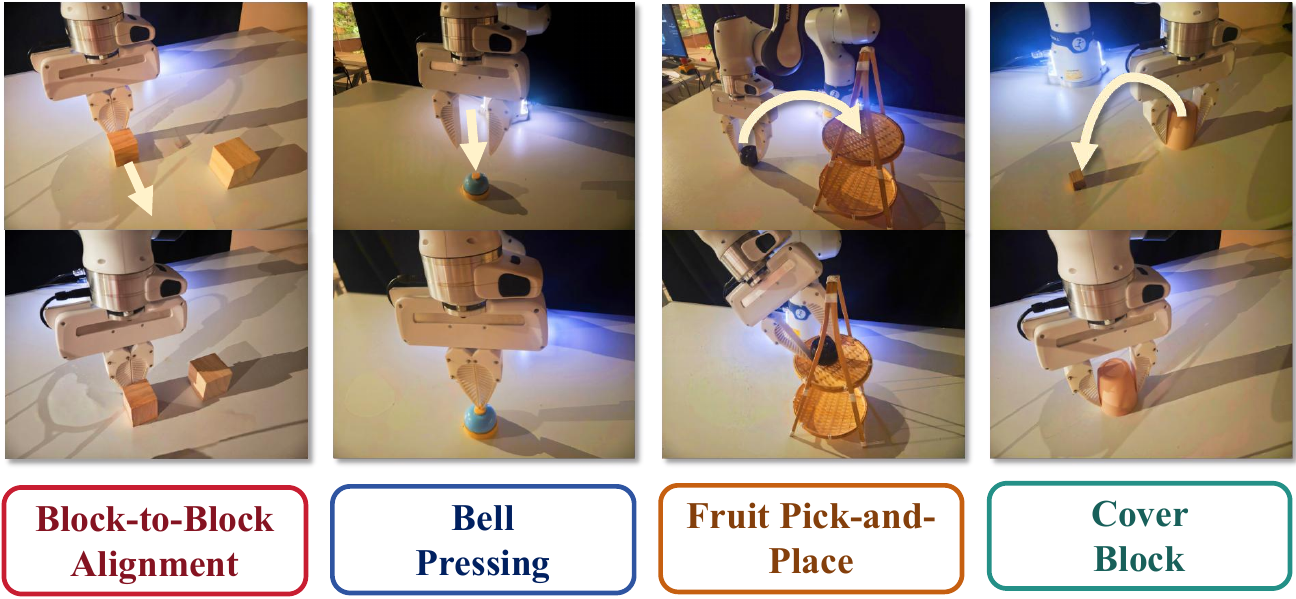}
    \caption{\textbf{Demonstration of Four Tasks in Real-World.}}
    \label{fig:demo}
\end{figure}

For all tasks (Fig.~\ref{fig:demo}), initial object poses are randomized within bounded perturbations to test robustness. For each task we collect \textbf{40 demonstrations} and report success rates over \textbf{20 trials}. A PointTransformer~\cite{pointtransformer} backbone serves as the visual encoder for all real-world experiments.

\subsection{Real-World Task Performance}
As shown in the \textcolor[HTML]{0070C0}{blue} block of Table~\ref{tab:real-world-in-out}, \textsc{AFRO} achieves the highest mean success rate of \textbf{0.70} across four tasks, outperforming DP3 (0.50), PointDif (0.53), FVP (0.34), and DynaMo-3D (0.48). The largest gains occur on \emph{Fruit Pick-and-Place} and \emph{Cover Block}, which require wide workspace motions (e.g., lifting fruit into an elevated basket or moving a cup over a target block). These results indicate that AFRO’s transition-centric 3D representations transfer reliably from simulation to a Franka setup under large spatial motions and real-world sensing noise.

\subsection{Large-Scale Out-of-Domain Pretraining}
We further evaluate scalability to large, out-of-domain data by pretraining on the RH20T~\cite{rh20t} real-world manipulation dataset. Specifically, we construct a Franka subset (RH20T\_Franka\_Simple) by reconstructing point clouds from camera intrinsics and depth, cropping to a fixed workspace, discarding the first 30 static frames, and sampling one frame every 20 from the first two scenes of over 140 tasks. On this corpus, we pretrain a shared PointTransformer encoder with \textbf{AFRO}, \textbf{DynaMo-3D}, \textbf{PointDif}, and \textbf{FVP}, then fine-tune it on four real-robot manipulation tasks with a diffusion action decoder. As shown in the \textcolor[HTML]{C00000}{red} block of Table~\ref{tab:real-world-in-out}, AFRO consistently benefits the most from large-scale out-of-domain pretraining, raising mean success from \textbf{70.0\%} to \textbf{84\%} and clearly outperforming all competing methods, indicating a stronger ability to effectively exploit heterogeneous real-world demonstrations for transferable 3D manipulation dynamics.

\begin{table}[t]
    \caption{\textbf{Objects generalization performance in two tasks.}}
    \centering
    \begin{minipage}{\linewidth}
        \centering
        \includegraphics[width=\linewidth]{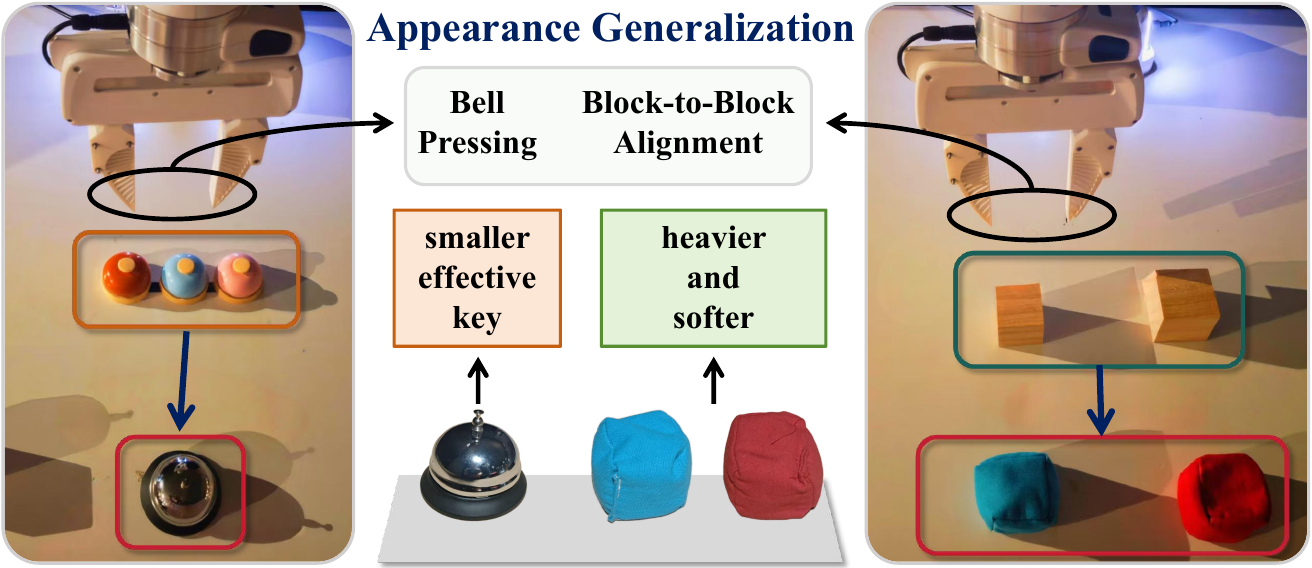}
    \end{minipage}
    \begin{minipage}{\linewidth}
        \centering
        \renewcommand{\arraystretch}{0.8}
        \small
        \setlength{\tabcolsep}{6pt}
        \begin{tabular}{lcccc}
            \toprule
            \multirow{2}{*}{\textbf{Method}} & \multicolumn{2}{c}{\textbf{Bell Pressing}} & \multicolumn{2}{c}{\textbf{B2B Alignment}} \\
            \cmidrule(lr){2-3} \cmidrule(lr){4-5}
            & \textbf{SR (\%)} & D\textbf{iff $\downarrow$} & \textbf{SR (\%)} & \textbf{Diff $\downarrow$} \\
            \midrule
            PointDif   & 55 $\rightarrow$ 25 & \textbf{-30} & 55 $\rightarrow$ 45 & \textbf{-10} \\
            FVP         & 40 $\rightarrow$ 5  & \textbf{-35} & 30 $\rightarrow$ 10 & \textbf{-20} \\
            DynaMo-3d   & 40 $\rightarrow$ 15 & \textbf{-25} & 50 $\rightarrow$ 35 & \textbf{-15} \\
            \rowcolor[HTML]{F9F9F9}
            \rowcolor[HTML]{F9F9F9}
            \textbf{AFRO (Ours)}        & \textbf{75 $\rightarrow$ 60} & \textbf{-15} & \textbf{70 $\rightarrow$ 65} & \textbf{-5} \\
            \bottomrule
        \end{tabular}
    \end{minipage}
    \label{tab:generalization}
\end{table}

\subsection{Generalization Evaluation}

\noindent
\textbf{Object Generalization.}
We test transfer to unseen objects in \emph{Bell Pressing} and \emph{Block-to-Block Alignment}. 
As shown in Table~\ref{tab:generalization}, all baselines drop substantially when moving from seen to unseen objects (e.g., up to 35 points for FVP on \emph{Bell Pressing}). 
\textsc{AFRO} achieves the highest success and the smallest gaps (75$\rightarrow$60 on \emph{BP}, 70$\rightarrow$65 on \emph{B2BA}), indicating its transition-centric 3D features capture task objectives beyond object appearance.

\noindent
\textbf{Cluttered-Scene Generalization.}
We further add distractors and clutter while keeping task definitions unchanged. 
Table~\ref{tab:cluttered} shows that all methods degrade, but AFRO remains stable, dropping only 5 points on both tasks (75$\rightarrow$70 for \emph{Bell Pressing}, 70$\rightarrow$65 for \emph{B2BA}), while PointDif and FVP suffer larger losses. 
This suggests AFRO’s 3D dynamics modeling is less sensitive to scene clutter, improving generalization across objects and environments.

\subsection{Ablation Study} As summarized in Table~\ref{tab:ablation}, removing any core component of AFRO consistently degrades performance on the four real-robot tasks, confirming the necessity of our overall design. Replacing the diffusion-based forward dynamics with a deterministic transformer (FDM: Transformer) lowers the mean success from \textbf{84.0\%} to \textbf{76.25\%}, highlighting the benefit of explicitly modeling multimodal futures in latent space. Dropping feature differencing or inverse-consistency supervision (w/o $\Delta \mathbf{z}$ Input, w/o Inv. Consistency) causes further drops, indicating that learning latent actions from representation changes and enforcing bidirectional consistency are both important to prevent feature leakage and stabilize training. Switching the pretraining objective from the variance-regularized VICReg loss to a plain MSE on features (\(\mathcal{L}_{\text{VICReg}} \rightarrow \mathcal{L}_{\text{MSE}}\)) leads to a large collapse in mean success (\textbf{57.50\%}), confirming that latent feature learning is prone to collapse without constraints on the feature distribution. Varying the latent action dimension (\(D_{\text{act}}=64,128\)) and the temporal interval used for frame sampling in pretraining (Interval $k=1,16$, vs.\ our default $k=4$) yields intermediate performance that never surpasses AFRO, suggesting that our chosen latent capacity and moderate frame interval strike a favorable balance between expressiveness and temporal coverage in practice.

\begin{table}[t]
    \caption{\textbf{Generalization Comparison: Cluttered Scenes.}}
    \centering
    \begin{minipage}{\linewidth}
        \centering
        \includegraphics[width=\linewidth]{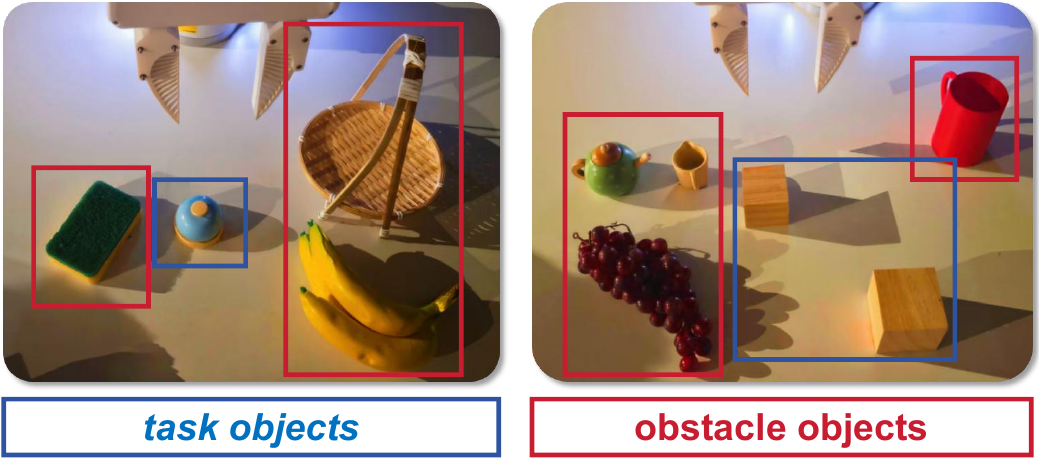}
    \end{minipage}
    \begin{minipage}{\linewidth}
        \centering
        \renewcommand{\arraystretch}{0.8}
        \small
        \setlength{\tabcolsep}{6pt}
        \begin{tabular}{lcccc}
            \toprule
            \multirow{2}{*}{\textbf{Method}} & \multicolumn{2}{c}{\textbf{Bell Pressing}} & \multicolumn{2}{c}{\textbf{B2B Alignment}} \\
            \cmidrule(lr){2-3} \cmidrule(lr){4-5}
            & \textbf{SR (\%)} & D\textbf{iff $\downarrow$} & \textbf{SR (\%)} & \textbf{Diff $\downarrow$} \\
            \midrule
            PointDif   & 55 $\rightarrow$ 40 & \textbf{-15} & 55 $\rightarrow$ 45 & \textbf{-10} \\
            FVP         & 40 $\rightarrow$ 25  & \textbf{-15} & 30 $\rightarrow$ 15 & \textbf{-15} \\
            DynaMo-3d   & 40 $\rightarrow$ 35 & \textbf{-5} & 50 $\rightarrow$ 40 & \textbf{-10} \\
            \rowcolor[HTML]{F9F9F9}
            \rowcolor[HTML]{F9F9F9}
            \textbf{AFRO (Ours)}        & \textbf{75 $\rightarrow$ 70} & \textbf{-5} & \textbf{70 $\rightarrow$ 65} & \textbf{-5} \\
            \bottomrule
        \end{tabular}
    \end{minipage}
    \label{tab:cluttered}
\end{table}

\begin{table}[t]
\caption{\textbf{Ablation on AFRO components on real-robot tasks.}}
\centering
\small
\setlength{\tabcolsep}{4pt}
\renewcommand{\arraystretch}{0.95}
\begin{tabular}{lccccc}
\toprule
                                   & FPP         & BP          & CB          & B2BA         & \textbf{Mean S.R.} \\ \midrule
FDM: Transformer                   & 70          & 80          & 80          & 75          & 76.25              \\
w/o $\Delta \mathbf{z}$ Input      & 65          & 80          & 75          & 80          & 75.00              \\
w/o Inv. Consistency               & 70          & 75          & 75          & 75          & 73.75              \\
$\mathcal{L}_{\text{VICReg}}$ $\rightarrow$ $\mathcal{L}_{\text{mse}}$    & 60          & 55          & 45          & 70          & 57.50              \\
$D_{\text{act}}=64$                & 70          & 90          & 80          & 80          & 80.00              \\
$D_{\text{act}}=128$               & 65          & 85          & 75          & 80          & 76.25              \\
Interval $k=1$                      & 75          & 85          & 80          & 85          & 81.25              \\
Interval $k=16$                     & 70          & 90          & 75          & 75          & 77.50              \\ \midrule
\rowcolor[HTML]{F9F9F9}
\textbf{AFRO (Ours)}               & \textbf{75} & \textbf{90} & \textbf{85} & \textbf{85} & \textbf{84.00}     \\
\bottomrule
\end{tabular}
\label{tab:ablation}
\end{table}

\section{Conclusion}
We introduced AFRO, an action-free 3D visual pre-training framework that learns dynamics-aware, manipulation-relevant representations in latent space by combining latent actions with diffusion-based forward dynamics, achieving strong performance and generalization on simulated and real-world manipulation tasks. Free from action labels, AFRO naturally scales to large unlabeled 3D robot interaction data, including data generated in simulation. A natural next step is to fuse AFRO with semantic priors from visual foundation models, targeting 3D representations that are both dynamics-aware and semantically grounded, further broadening open-world manipulation capabilities.

{
    \small
    \bibliographystyle{ieeenat_fullname}
    \bibliography{main}

@String(CVPR= {IEEE Conf. Comput. Vis. Pattern Recog.})

@String(ICCV= {Int. Conf. Comput. Vis.})

@String(ECCV= {Eur. Conf. Comput. Vis.})

@String(ICLR = {Int. Conf. Learn. Represent.})

@String(CVPR  = {CVPR})

@String(ICCV  = {ICCV})

@String(ECCV  = {ECCV})

@String(ICLR  = {ICLR})

@inproceedings{adroit_ref,
  author       = {Aravind Rajeswaran and Vikash Kumar and Abhishek Gupta and Giulia Vezzani and John Schulman and Emanuel Todorov and Sergey Levine},
  editor       = {Hadas Kress{-}Gazit and Siddhartha S. Srinivasa and Tom Howard and Nikolay Atanasov},
  title        = {Learning Complex Dexterous Manipulation with Deep Reinforcement Learning and Demonstrations},
  booktitle    = {Robotics: Science and Systems (RSS)},
  year         = {2018},
}

@inproceedings{metaworld_ref,
  title={Meta-world: A benchmark and evaluation for multi-task and meta reinforcement learning},
  author={Yu, Tianhe and Quillen, Deirdre and He, Zhanpeng and Julian, Ryan and Hausman, Karol and Finn, Chelsea and Levine, Sergey},
  booktitle={Conference on robot learning (CoRL)},
  pages={1094--1100},
  year={2020},
  organization={PMLR}
}

@inproceedings{dp3_ref,
title={3D Diffusion Policy: Generalizable Visuomotor Policy Learning via Simple 3D Representations},
author={Yanjie Ze and Gu Zhang and Kangning Zhang and Chenyuan Hu and Muhan Wang and Huazhe Xu},
booktitle={2nd Workshop on Dexterous Manipulation: Design, Perception and Control (RSS)},
year={2024},
url={https://openreview.net/forum?id=KwwJuZIBXH}
}

@article{vrl3_ref,
  title={Vrl3: A data-driven framework for visual deep reinforcement learning},
  author={Wang, Che and Luo, Xufang and Ross, Keith and Li, Dongsheng},
  journal={Advances in Neural Information Processing Systems (NeurIPS)},
  volume={35},
  pages={32974--32988},
  year={2022}
}

@inproceedings{fvp_ref,
  title={4D Visual Pre-training for Robot Learning},
  author={Hou, Chengkai and Ze, Yanjie and Fu, Yankai and Gao, Zeyu and Hu, Songbo and Yu, Yue and Zhang, Shanghang and Xu, Huazhe},
  booktitle={Proceedings of the IEEE/CVF International Conference on Computer Vision (ICCV)},
  pages={8451--8461},
  year={2025}
}

@inproceedings{pointnet_ref,
  title={Pointnet: Deep learning on point sets for 3d classification and segmentation},
  author={Qi, Charles R and Su, Hao and Mo, Kaichun and Guibas, Leonidas J},
  booktitle={Proceedings of the IEEE conference on computer vision and pattern recognition (CVPR)},
  pages={652--660},
  year={2017}
}

@inproceedings{clip_ref,
  title={Learning transferable visual models from natural language supervision},
  author={Radford, Alec and Kim, Jong Wook and Hallacy, Chris and Ramesh, Aditya and Goh, Gabriel and Agarwal, Sandhini and Sastry, Girish and Askell, Amanda and Mishkin, Pamela and Clark, Jack and others},
  booktitle={International conference on machine learning (ICML)},
  pages={8748--8763},
  year={2021},
  organization={PmLR}
}

@article{dinov2_ref,
  title={Dinov2: Learning robust visual features without supervision},
  author={Oquab, Maxime and Darcet, Timoth{\'e}e and Moutakanni, Th{\'e}o and Vo, Huy and Szafraniec, Marc and Khalidov, Vasil and Fernandez, Pierre and Haziza, Daniel and Massa, Francisco and El-Nouby, Alaaeldin and others},
  journal={arXiv preprint arXiv:2304.07193},
  year={2023}
}

@inproceedings{pointmae_ref,
  title={Masked Autoencoders for Point Cloud Self-supervised Learning},
  author={Pang, Yatian and Wang, Wenxiao and Tay, Francis EH and Liu, Wei and Tian, Yonghong and Yuan, Li},
  booktitle={European Conference on Computer Vision (ECCV)},
  pages={604--621},
  year={2022}
}

@inproceedings{pointdiff_ref,
  title={Point cloud pre-training with diffusion models},
  author={Zheng, Xiao and Huang, Xiaoshui and Mei, Guofeng and Hou, Yuenan and Lyu, Zhaoyang and Dai, Bo and Ouyang, Wanli and Gong, Yongshun},
  booktitle={Proceedings of the IEEE/CVF Conference on Computer Vision and Pattern Recognition (CVPR)},
  pages={22935--22945},
  year={2024}
}

@article{dynamo_ref,
  title={Dynamo: In-domain dynamics pretraining for visuo-motor control},
  author={Cui, Zichen J and Pan, Hengkai and Iyer, Aadhithya and Haldar, Siddhant and Pinto, Lerrel},
  journal={Advances in Neural Information Processing Systems},
  volume={37},
  pages={33933--33961},
  year={2024}
}

@article{mvp_ref,
  title={Masked Visual Pre-training for Motor Control},
  author={Tete Xiao and Ilija Radosavovic and Trevor Darrell and Jitendra Malik},
  journal={arXiv preprint arXiv:2203.06173},
  year={2022}
}

@inproceedings{r3m_ref,
  title={R3M: A Universal Visual Representation for Robot Manipulation},
  author={Nair, Suraj and Rajeswaran, Aravind and Kumar, Vikash and Finn, Chelsea and Gupta, Abhinav},
  booktitle={Conference on Robot Learning (CoRL)},
  pages={892--909},
  year={2023},
  organization={PMLR}
}

@inproceedings{vip_ref,
  title={VIP: Towards Universal Visual Reward and Representation via Value-Implicit Pre-Training},
  author={Ma, Yecheng Jason and Sodhani, Shagun and Jayaraman, Dinesh and Bastani, Osbert and Kumar, Vikash and Zhang, Amy},
  booktitle={The Eleventh International Conference on Learning Representations (ICLR)},
  year={2023},
}

@inproceedings{3d-mvp,
  title={3D-MVP: 3D Multiview Pretraining for Manipulation},
  author={Qian, Shengyi and Mo, Kaichun and Blukis, Valts and Fouhey, David F and Fox, Dieter and Goyal, Ankit},
  booktitle={Proceedings of the Computer Vision and Pattern Recognition Conference (CVPR)},
  pages={22530--22539},
  year={2025}
}

@inproceedings{sugar,
  title={Sugar: Pre-training 3d visual representations for robotics},
  author={Chen, Shizhe and Garcia, Ricardo and Laptev, Ivan and Schmid, Cordelia},
  booktitle={Proceedings of the IEEE/CVF Conference on Computer Vision and Pattern Recognition (CVPR)},
  pages={18049--18060},
  year={2024}
}

@article{clam,
  title={Clam: Continuous latent action models for robot learning from unlabeled demonstrations},
  author={Liang, Anthony and Czempin, Pavel and Hong, Matthew and Zhou, Yutai and Biyik, Erdem and Tu, Stephen},
  journal={arXiv preprint arXiv:2505.04999},
  year={2025}
}

@article{como,
  title={CoMo: Learning Continuous Latent Motion from Internet Videos for Scalable Robot Learning},
  author={Yang, Jiange and Shi, Yansong and Zhu, Haoyi and Liu, Mingyu and Ma, Kaijing and Wang, Yating and Wu, Gangshan and He, Tong and Wang, Limin},
  journal={arXiv preprint arXiv:2505.17006},
  year={2025}
}

@article{lapa,
  title={Latent action pretraining from videos},
  author={Ye, Seonghyeon and Jang, Joel and Jeon, Byeongguk and Joo, Sejune and Yang, Jianwei and Peng, Baolin and Mandlekar, Ajay and Tan, Reuben and Chao, Yu-Wei and Lin, Bill Yuchen and others},
  journal={arXiv preprint arXiv:2410.11758},
  year={2024}
}

@article{latent-world,
  title={Latent Action Pretraining Through World Modeling},
  author={Tharwat, Bahey and Nasser, Yara and Abouzeid, Ali and Reid, Ian},
  journal={arXiv preprint arXiv:2509.18428},
  year={2025}
}

@article{villa,
  title={Villa-x: enhancing latent action modeling in vision-language-action models},
  author={Chen, Xiaoyu and Wei, Hangxing and Zhang, Pushi and Zhang, Chuheng and Wang, Kaixin and Guo, Yanjiang and Yang, Rushuai and Wang, Yucen and Xiao, Xinquan and Zhao, Li and others},
  journal={arXiv preprint arXiv:2507.23682},
  year={2025}
}

@article{univla,
  title={Univla: Learning to act anywhere with task-centric latent actions},
  author={Bu, Qingwen and Yang, Yanting and Cai, Jisong and Gao, Shenyuan and Ren, Guanghui and Yao, Maoqing and Luo, Ping and Li, Hongyang},
  journal={arXiv preprint arXiv:2505.06111},
  year={2025}
}

@inproceedings{plas,
  title={Plas: Latent action space for offline reinforcement learning},
  author={Zhou, Wenxuan and Bajracharya, Sujay and Held, David},
  booktitle={Conference on Robot Learning (CoRL)},
  pages={1719--1735},
  year={2021},
  organization={PMLR}
}

@article{taco,
  title={\texttt{TACO}: Temporal Latent Action-Driven Contrastive Loss for Visual Reinforcement Learning},
  author={Zheng, Ruijie and Wang, Xiyao and Sun, Yanchao and Ma, Shuang and Zhao, Jieyu and Xu, Huazhe and Daum{\'e} III, Hal and Huang, Furong},
  journal={Advances in Neural Information Processing Systems},
  volume={36},
  pages={48203--48225},
  year={2023}
}

@article{jif,
  title={Train Robots in a JIF: Joint Inverse and Forward Dynamics with Human and Robot Demonstrations},
  author={Khandate, Gagan and Wang, Boxuan and Park, Sarah and Ni, Weizhe and Palacios, Joaquin and Lampo, Kathyrn and Wu, Philippe and Ho, Rosh and Chang, Eric and Ciocarlie, Matei},
  journal={arXiv preprint arXiv:2503.12297},
  year={2025}
}

@article{AMPLIFY,
  title={AMPLIFY: Actionless Motion Priors for Robot Learning from Videos},
  author={Collins, Jeremy A and Cheng, Lor{\'a}nd and Aneja, Kunal and Wilcox, Albert and Joffe, Benjamin and Garg, Animesh},
  journal={arXiv preprint arXiv:2506.14198},
  year={2025}
}

@article{latent-supervision,
  title={Latent action learning requires supervision in the presence of distractors},
  author={Nikulin, Alexander and Zisman, Ilya and Tarasov, Denis and Lyubaykin, Nikita and Polubarov, Andrei and Kiselev, Igor and Kurenkov, Vladislav},
  journal={arXiv preprint arXiv:2502.00379},
  year={2025}
}

@article{latent-behavior,
  title={Behavior generation with latent actions},
  author={Lee, Seungjae and Wang, Yibin and Etukuru, Haritheja and Kim, H Jin and Shafiullah, Nur Muhammad Mahi and Pinto, Lerrel},
  journal={arXiv preprint arXiv:2403.03181},
  year={2024}
}

@article{adaworld,
  title={Adaworld: Learning adaptable world models with latent actions},
  author={Gao, Shenyuan and Zhou, Siyuan and Du, Yilun and Zhang, Jun and Gan, Chuang},
  journal={arXiv preprint arXiv:2503.18938},
  year={2025}
}

@article{hrp,
  title={Hrp: Human affordances for robotic pre-training},
  author={Srirama, Mohan Kumar and Dasari, Sudeep and Bahl, Shikhar and Gupta, Abhinav},
  journal={arXiv preprint arXiv:2407.18911},
  year={2024}
}

@inproceedings{sensorimotor,
  title={Robot learning with sensorimotor pre-training},
  author={Radosavovic, Ilija and Shi, Baifeng and Fu, Letian and Goldberg, Ken and Darrell, Trevor and Malik, Jitendra},
  booktitle={Conference on Robot Learning (CoRL)},
  pages={683--693},
  year={2023},
  organization={PMLR}
}

@article{RPR,
  title={Robots pre-train robots: Manipulation-centric robotic representation from large-scale robot datasets},
  author={Jiang, Guangqi and Sun, Yifei and Huang, Tao and Li, Huanyu and Liang, Yongyuan and Xu, Huazhe},
  journal={arXiv preprint arXiv:2410.22325},
  year={2024}
}

@inproceedings{unbiased-pre-training,
  title={An unbiased look at datasets for visuo-motor pre-training},
  author={Dasari, Sudeep and Srirama, Mohan Kumar and Jain, Unnat and Gupta, Abhinav},
  booktitle={Conference on Robot Learning (CoRL)},
  pages={1183--1198},
  year={2023},
  organization={PMLR}
}

@inproceedings{real-pre,
  title={Real-world robot learning with masked visual pre-training},
  author={Radosavovic, Ilija and Xiao, Tete and James, Stephen and Abbeel, Pieter and Malik, Jitendra and Darrell, Trevor},
  booktitle={Conference on Robot Learning (CoRL)},
  pages={416--426},
  year={2023},
  organization={PMLR}
}

@article{points,
  title={Point cloud matters: Rethinking the impact of different observation spaces on robot learning},
  author={Zhu, Haoyi and Wang, Yating and Huang, Di and Ye, Weicai and Ouyang, Wanli and He, Tong},
  journal={Advances in Neural Information Processing Systems},
  volume={37},
  pages={77799--77830},
  year={2024}
}

@article{whole,
  title={Whole-Body Coordination for Dynamic Object Grasping with Legged Manipulators},
  author={Liang, Qiwei and Cai, Boyang and He, Rongyi and Li, Hui and Teng, Tao and Duan, Haihan and Huang, Changxin and Zeng, Runhao},
  journal={arXiv preprint arXiv:2508.08328},
  year={2025}
}

@inproceedings{rvt,
  title={Rvt: Robotic view transformer for 3d object manipulation},
  author={Goyal, Ankit and Xu, Jie and Guo, Yijie and Blukis, Valts and Chao, Yu-Wei and Fox, Dieter},
  booktitle={Conference on Robot Learning (CoRL)},
  pages={694--710},
  year={2023},
  organization={PMLR}
}

@article{lift3d,
  title={Lift3d foundation policy: Lifting 2d large-scale pretrained models for robust 3d robotic manipulation},
  author={Jia, Yueru and Liu, Jiaming and Chen, Sixiang and Gu, Chenyang and Wang, Zhilue and Luo, Longzan and Lee, Lily and Wang, Pengwei and Wang, Zhongyuan and Zhang, Renrui and others},
  journal={arXiv preprint arXiv:2411.18623},
  year={2024}
}

@inproceedings{strl,
  title={Spatio-temporal self-supervised representation learning for 3d point clouds},
  author={Huang, Siyuan and Xie, Yichen and Zhu, Song-Chun and Zhu, Yixin},
  booktitle={Proceedings of the IEEE/CVF international conference on computer vision (ICCV)},
  pages={6535--6545},
  year={2021}
}

@inproceedings{pointbert,
  title={Point-bert: Pre-training 3d point cloud transformers with masked point modeling},
  author={Yu, Xumin and Tang, Lulu and Rao, Yongming and Huang, Tiejun and Zhou, Jie and Lu, Jiwen},
  booktitle={Proceedings of the IEEE/CVF conference on computer vision and pattern recognition (CVPR)},
  pages={19313--19322},
  year={2022}
}

@inproceedings{pointcontrast,
  title={Pointcontrast: Unsupervised pre-training for 3d point cloud understanding},
  author={Xie, Saining and Gu, Jiatao and Guo, Demi and Qi, Charles R and Guibas, Leonidas and Litany, Or},
  booktitle={European conference on computer vision (ECCV)},
  pages={574--591},
  year={2020},
  organization={Springer}
}

@article{robomind,
  title={Robomind: Benchmark on multi-embodiment intelligence normative data for robot manipulation},
  author={Wu, Kun and Hou, Chengkai and Liu, Jiaming and Che, Zhengping and Ju, Xiaozhu and Yang, Zhuqin and Li, Meng and Zhao, Yinuo and Xu, Zhiyuan and Yang, Guang and others},
  journal={arXiv preprint arXiv:2412.13877},
  year={2024}
}

@article{rh20t,
  title={Rh20t: A comprehensive robotic dataset for learning diverse skills in one-shot},
  author={Fang, Hao-Shu and Fang, Hongjie and Tang, Zhenyu and Liu, Jirong and Wang, Chenxi and Wang, Junbo and Zhu, Haoyi and Lu, Cewu},
  journal={arXiv preprint arXiv:2307.00595},
  year={2023}
}

@article{v-jepa2,
  title={V-jepa 2: Self-supervised video models enable understanding, prediction and planning},
  author={Assran, Mido and Bardes, Adrien and Fan, David and Garrido, Quentin and Howes, Russell and Muckley, Matthew and Rizvi, Ammar and Roberts, Claire and Sinha, Koustuv and Zholus, Artem and others},
  journal={arXiv preprint arXiv:2506.09985},
  year={2025}
}

@article{vicreg,
  title={Vicreg: Variance-invariance-covariance regularization for self-supervised learning},
  author={Bardes, Adrien and Ponce, Jean and LeCun, Yann},
  journal={arXiv preprint arXiv:2105.04906},
  year={2021}
}

@inproceedings{allshire2021laser,
  title={Laser: Learning a latent action space for efficient reinforcement learning},
  author={Allshire, Arthur and Mart{\'\i}n-Mart{\'\i}n, Roberto and Lin, Charles and Manuel, Shawn and Savarese, Silvio and Garg, Animesh},
  booktitle={2021 IEEE International Conference on Robotics and Automation (ICRA)},
  pages={6650--6656},
  year={2021},
  organization={IEEE}
}

@article{embodiedmae,
  title={EmbodiedMAE: A Unified 3D Multi-Modal Representation for Robot Manipulation},
  author={Dong, Zibin and Ni, Fei and Yuan, Yifu and Li, Yinchuan and Hao, Jianye},
  journal={arXiv preprint arXiv:2505.10105},
  year={2025}
}

@inproceedings{pointtransformer,
  title={Point transformer},
  author={Zhao, Hengshuang and Jiang, Li and Jia, Jiaya and Torr, Philip HS and Koltun, Vladlen},
  booktitle={Proceedings of the IEEE/CVF international conference on computer vision ICCV},
  pages={16259--16268},
  year={2021}
}

@inproceedings{ar_vrm,
  title={AR-VRM: Imitating Human Motions for Visual Robot Manipulation with Analogical Reasoning},
  author={Yang, Dejie and Zhao, Zijing and Liu, Yang},
  booktitle={Proceedings of the IEEE/CVF International Conference on Computer Vision (ICCV)},
  pages={6818--6827},
  year={2025}
}

@inproceedings{dit,
  title={Scalable diffusion models with transformers},
  author={Peebles, William and Xie, Saining},
  booktitle={Proceedings of the IEEE/CVF international conference on computer vision},
  pages={4195--4205},
  year={2023}
}

@inproceedings{exploring,
  title={Exploring visual pre-training for robot manipulation: Datasets, models and methods},
  author={Jing, Ya and Zhu, Xuelin and Liu, Xingbin and Sima, Qie and Yang, Taozheng and Feng, Yunhai and Kong, Tao},
  booktitle={2023 IEEE/RSJ International Conference on Intelligent Robots and Systems (IROS)},
  pages={11390--11395},
  year={2023},
}

@inproceedings{learningact,
  title={Learning to see before learning to act: Visual pre-training for manipulation},
  author={Yen-Chen, Lin and Zeng, Andy and Song, Shuran and Isola, Phillip and Lin, Tsung-Yi},
  booktitle={2020 IEEE International Conference on Robotics and Automation (ICRA)},
  pages={7286--7293},
  year={2020},
  organization={IEEE}
}

@article{dexsingrasp,
  title={DexSinGrasp: Learning a Unified Policy for Dexterous Object Singulation and Grasping in Cluttered Environments},
  author={Xu, Lixin and Liu, Zixuan and Gui, Zhewei and Guo, Jingxiang and Jiang, Zeyu and Xu, Zhixuan and Gao, Chongkai and Shao, Lin},
  journal={arXiv preprint arXiv:2504.04516},
  year={2025}
}

@article{zheng2024gaussiangrasper,
  title={Gaussiangrasper: 3d language gaussian splatting for open-vocabulary robotic grasping},
  author={Zheng, Yuhang and Chen, Xiangyu and Zheng, Yupeng and Gu, Songen and Yang, Runyi and Jin, Bu and Li, Pengfei and Zhong, Chengliang and Wang, Zengmao and Liu, Lina and others},
  journal={IEEE Robotics and Automation Letters},
  year={2024},
  publisher={IEEE}
}

@InProceedings{Wang_2025_ICCV,
    author    = {Wang, Taowen and Han, Cheng and Liang, James and Yang, Wenhao and Liu, Dongfang and Zhang, Luna Xinyu and Wang, Qifan and Luo, Jiebo and Tang, Ruixiang},
    title     = {Exploring the Adversarial Vulnerabilities of Vision-Language-Action Models in Robotics},
    booktitle = {Proceedings of the IEEE/CVF International Conference on Computer Vision (ICCV)},
    month     = {October},
    year      = {2025},
    pages     = {6948-6958}
}

@article{wu2024tars,
  title={TARS: Tactile Affordance in Robot Synesthesia for Dexterous Manipulation},
  author={Wu, Qiwei and Wang, Haidong and Zhou, Jiayu and Xiong, Xiaogang and Lou, Yunjiang},
  journal={IEEE Robotics and Automation Letters},
  year={2024},
  publisher={IEEE}
}

@inproceedings{wu2024rttf,
  title={RTTF: Rapid Tactile Transfer Framework for Contact-Rich Manipulation Tasks},
  author={Wu, Qiwei and Peng, Xuanbin and Zhou, Jiayu and Sun, Zhuoran and Xiong, Xiaogang and Lou, Yunjiang},
  booktitle={2024 IEEE/RSJ International Conference on Intelligent Robots and Systems (IROS)},
  pages={2913--2920},
  year={2024},
  organization={IEEE}
}

@article{zhou2025gentle,
  title={Gentle Manipulation Policy Learning via Demonstrations from VLM Planned Atomic Skills},
  author={Zhou, Jiayu and Wu, Qiwei and Li, Jian and Chen, Zhe and Xiong, Xiaogang and Xu, Renjing},
  journal={arXiv preprint arXiv:2511.05855},
  year={2025}
}

@article{gervet2023act3d,
  title={Act3d: 3d feature field transformers for multi-task robotic manipulation},
  author={Gervet, Theophile and Xian, Zhou and Gkanatsios, Nikolaos and Fragkiadaki, Katerina},
  journal={arXiv preprint arXiv:2306.17817},
  year={2023}
}

@inproceedings{shridhar2023perceiver,
  title={Perceiver-actor: A multi-task transformer for robotic manipulation},
  author={Shridhar, Mohit and Manuelli, Lucas and Fox, Dieter},
  booktitle={Conference on Robot Learning},
  pages={785--799},
  year={2023},
  organization={PMLR}
}

@inproceedings{forcefb,
  title={Kinematic multi-robot manipulation with no communication using force feedback},
  author={Wang, Zijian and Schwager, Mac},
  booktitle={2016 IEEE international conference on robotics and automation (ICRA)},
  pages={427--432},
  year={2016},
  organization={IEEE}
}

@article{kim2024openvla,
  title={Openvla: An open-source vision-language-action model},
  author={Kim, Moo Jin and Pertsch, Karl and Karamcheti, Siddharth and Xiao, Ted and Balakrishna, Ashwin and Nair, Suraj and Rafailov, Rafael and Foster, Ethan and Lam, Grace and Sanketi, Pannag and others},
  journal={arXiv preprint arXiv:2406.09246},
  year={2024}
}

@article{huang2025tactile,
  title={Tactile-VLA: Unlocking Vision-Language-Action Model's Physical Knowledge for Tactile Generalization},
  author={Huang, Jialei and Wang, Shuo and Lin, Fanqi and Hu, Yihang and Wen, Chuan and Gao, Yang},
  journal={arXiv preprint arXiv:2507.09160},
  year={2025}
}

@article{chi2025diffusion,
  title={Diffusion policy: Visuomotor policy learning via action diffusion},
  author={Chi, Cheng and Xu, Zhenjia and Feng, Siyuan and Cousineau, Eric and Du, Yilun and Burchfiel, Benjamin and Tedrake, Russ and Song, Shuran},
  journal={The International Journal of Robotics Research},
  volume={44},
  number={10-11},
  pages={1684--1704},
  year={2025},
}

@inproceedings{wang2025vggt,
  title={Vggt: Visual geometry grounded transformer},
  author={Wang, Jianyuan and Chen, Minghao and Karaev, Nikita and Vedaldi, Andrea and Rupprecht, Christian and Novotny, David},
  booktitle={Proceedings of the Computer Vision and Pattern Recognition Conference},
  pages={5294--5306},
  year={2025}
}

@article{ye2026mind,
  title={MIND: Benchmarking Memory Consistency and Action Control in World Models},
  author={Ye, Yixuan and Lu, Xuanyu and Jiang, Yuxin and Gu, Yuchao and Zhao, Rui and Liang, Qiwei and Pan, Jiachun and Zhang, Fengda and Wu, Weijia and Wang, Alex Jinpeng},
  journal={arXiv preprint arXiv:2602.08025},
  year={2026}
}
}

\newpage
\appendix
\section{Abstract}
In this appendix, we complement the main paper with additional analyses and results for AFRO.
We first situate our approach within the broader literature on robotic manipulation.
We then provide qualitative visualizations of AFRO rollouts on four real-world Franka tasks using point-cloud observations, as well as the complete per-task results on all Adroit and MetaWorld benchmarks.
Next, we describe in more detail how we process the large-scale RH20T dataset to obtain point-cloud inputs for pre-training.
Finally, we discuss current limitations of AFRO and outline several promising directions for scaling, enriching semantics, incorporating multi-view dynamics, and improving latent-space learning objectives.

\begin{figure*}[!t]
    \centering
    \includegraphics[width=\textwidth]{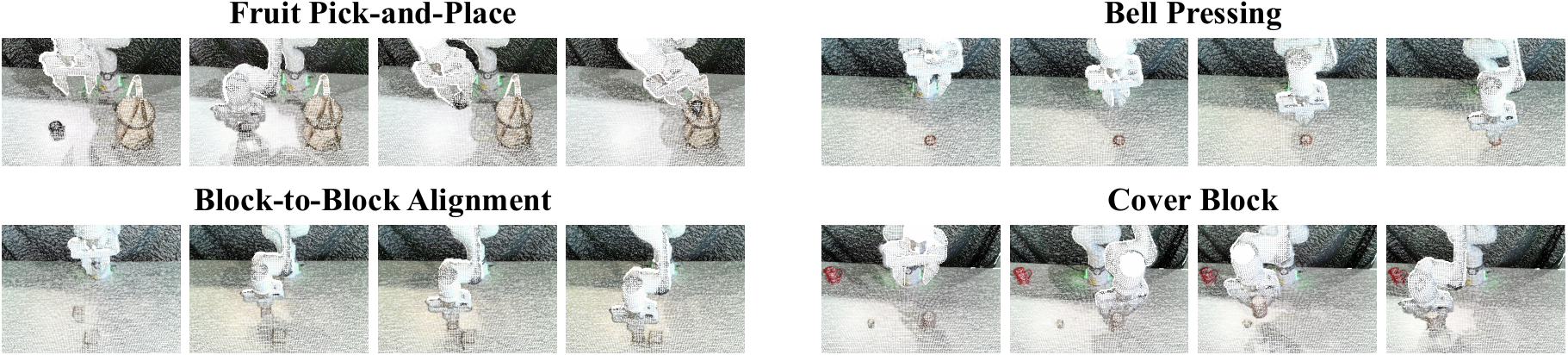}
    \caption{\textbf{Point-cloud rollouts for four real-world tasks.}
    We visualize the 3D point clouds captured by the top-down RealSense L515 depth camera for four representative manipulation tasks on the Franka platform: Fruit Pick-and-Place, Bell Pressing, Block-to-Block Alignment, and Cover Block (from left to right and top to bottom).
    Each block shows four temporally ordered point clouds from a successful AFRO rollout, illustrating the evolution from the initial configuration, through approach and contact, to task completion.
    Across tasks, the manipulated objects (fruit and basket, bell, blocks, cup and target block) undergo large spatial motion while the surrounding table and robot geometry remain largely static.
    These visualizations highlight that AFRO is trained directly on such raw point clouds and must learn dynamics-aware representations that are sensitive to object motion and interaction, yet robust to background clutter and viewpoint changes.}
    \label{fig:details_visualization}
\end{figure*}

\section{Additional Related Work}
\subsection{Robotics Manipulation.}
Robotic manipulation aims to transform object states in the physical world, from basic grasping\cite{zheng2024gaussiangrasper,dexsingrasp} to long-horizon, contact-rich~\cite{wu2024rttf} tasks. 
Early systems decomposed perception, grasp planning, and motion planning, while recent work learns end-to-end policies from rich sensory inputs such as RGB(-D)~\cite{chi2025diffusion}, tactile~\cite{wu2024tars,zhou2025gentle,huang2025tactile}, and force feedback~\cite{forcefb}, often with additional safety constraints~\cite{Wang_2025_ICCV} or language conditioning. 
Vision–language–action models~\cite{kim2024openvla} further condition policies on natural-language goals but are still mostly built on 2D image encoders, which struggle with occlusions and precise 6-DoF reasoning. 
To overcome these limitations, recent approaches use 3D-aware policies that operate on depth, multi-view RGB-D, or point clouds, such as methods that infer volumetric features or explicit 3D action maps~\cite{gervet2023act3d,rvt,shridhar2023perceiver}. 
Our work is complementary: instead of introducing another policy head, AFRO provides a 3D dynamics-aware representation that can be plugged into existing diffusion-based manipulation policies with minimal changes.

\section{Additional Real-World Visualizations}
\label{sec:additional_real_visual}
The main paper evaluates AFRO on four real-robot tasks that span non-prehensile pushing, precise contact interaction, and long-horizon pick-and-place motions using a Franka Emika arm and a top-down depth sensor.%
In Fruit Pick-and-Place, the robot must reach for a fruit at a randomized pose and place it into a distant basket; in Bell Pressing, it must localize a small bell and press it with sufficient accuracy to trigger the mechanism; in Block-to-Block Alignment, it pushes a movable block until its edge is aligned with a fixed reference block; and in Cover Block, it lifts a cup and places it so that the cup fully covers a target block.
In all cases, object positions and orientations are randomized within bounded ranges to test robustness.

Figure~\ref{fig:details_visualization} provides additional qualitative insight into how these tasks appear in the point-cloud observation space.
For each task we render four representative frames, showing the pre-contact configuration, the approach of the end-effector, the main interaction phase (pressing, pushing, or grasping), and the final configuration after successful completion.
The sequences illustrate the substantial 3D motion, self-occlusion, and depth noise present in real scenes, as well as the diversity of object geometries (e.g., thin bell handle versus bulky fruit or cup).
AFRO is pre-trained and fine-tuned directly on such sequences, and the strong real-world performance reported in the main paper indicates that its latent dynamics modeling can extract task-relevant structure from these raw point-cloud trajectories.

\section{Complete MetaWorld Results}
\label{sec:complete_metaworld}

Table~\ref{tab: complete metaworld results} reports per-task success rates on all 16 simulated manipulation tasks used in our study: two Adroit hand tasks (Door, Pen) and fourteen MetaWorld tasks spanning four difficulty levels (Easy, Medium, Hard, Very Hard).
For each method, we train a diffusion policy on top of the corresponding visual encoder using 100 expert trajectories for each Adroit task and 25 trajectories for each MetaWorld task, and evaluate success over 50 rollouts every 10 epochs; the table shows the best success rate over multiple runs under this shared protocol.
Compared to all baselines, AFRO attains the highest mean performance and achieves the best result on the majority of tasks, ranking second only on Bin Picking, Coffee Pull, and Soccer.
These fine-grained results complement the aggregated comparisons in the main paper and confirm that AFRO consistently improves over both 2D and 3D pre-training baselines across a broad range of contact-rich skills.

\begin{table*}[t]
\caption{\textbf{Comparison across 16 simulated manipulation tasks.}
Success rates (\%) for AFRO and competing methods on two Adroit tasks and fourteen MetaWorld tasks grouped by difficulty.
Best results are highlighted in \colorbox[HTML]{ffcccb}{red}, and second-best results are highlighted in \colorbox[HTML]{ffe5cc}{orange}.
All methods use the same diffusion-policy architecture and training protocol: policies are trained from 100 expert trajectories on Adroit and 25 trajectories on each MetaWorld task, and evaluated over 50 rollouts every 10 epochs; we report the best success rate over multiple runs.}
\centering
\resizebox{\textwidth}{!}{
\setlength{\tabcolsep}{4pt}
\renewcommand{\arraystretch}{1.1}
\begin{tabular}{l|ccccc|ccc}
\toprule[2pt]
         & \multicolumn{2}{c|}{\textbf{Adroit}}    & \multicolumn{3}{c|}{\textbf{MetaWorld (Easy)}}         & \multicolumn{3}{c}{\textbf{MetaWorld (Medium)}}      \\
\textbf{Method}  & Door      & \multicolumn{1}{c|}{Pen} & Dial Turn                  & Handle Press       & Peg Unplug Side  & Bin Picking & Coffee Pull & Peg Insert Side \\ 
\midrule
CILP      & 61        & \multicolumn{1}{c|}{\cellcolor[HTML]{ffcccb}84}  & 0 & 88  & 0  & 0 & 72  & 0               \\
DINOv2    & 76        & \multicolumn{1}{c|}{\cellcolor[HTML]{ffcccb}84}  & 6 & \cellcolor[HTML]{ffe5cc}90 & 6 & 0 & 58 & 0               \\
PointNet  & \cellcolor[HTML]{ffe5cc}80        & \multicolumn{1}{c|}{72}  & 56 & \cellcolor[HTML]{ffe5cc}90 & 68  & \cellcolor[HTML]{ffcccb}62     & 66       & 54              \\
PointMAE  & 64        & \multicolumn{1}{c|}{76}  & 60       & 88        & 84    & 16   & \cellcolor[HTML]{ffcccb}94    & 82              \\
PointDif  & 76        & \multicolumn{1}{c|}{78}  & \cellcolor[HTML]{ffe5cc}76    & \cellcolor[HTML]{ffe5cc}90   & 84  & 18   & 90    & \cellcolor[HTML]{ffe5cc}84              \\
Dynamo    & 76        & \multicolumn{1}{c|}{68}  & 18      & 84     & 24   & 18    & 34  & 14              \\
Dynamo-3D & 73        & \multicolumn{1}{c|}{76}  & 70    & \cellcolor[HTML]{ffcccb}98  & \cellcolor[HTML]{ffe5cc}86   & 18          & 68          & 76              \\
FVP       & 66        & \multicolumn{1}{c|}{76}  & 72    & 88   & 80    & 16   & 66   & 70              \\
DP3       & 70        & \multicolumn{1}{c|}{\cellcolor[HTML]{ffe5cc}80}  & \cellcolor[HTML]{ffe5cc}76  & 86    & \cellcolor[HTML]{ffe5cc}86  & 18   & 90   & 74              \\
\textbf{AFRO (Ours)}   & \cellcolor[HTML]{ffcccb}82   & \multicolumn{1}{c|}{\cellcolor[HTML]{ffcccb}84}  & \cellcolor[HTML]{ffcccb}78                 & \cellcolor[HTML]{ffcccb}98                 & \cellcolor[HTML]{ffcccb}88               & \cellcolor[HTML]{ffe5cc}20          & \cellcolor[HTML]{ffe5cc}92          & \cellcolor[HTML]{ffcccb}92              \\ 
\midrule[2pt]
         & \multicolumn{3}{c|}{\textbf{MetaWorld (Medium)}}       & \multicolumn{2}{c|}{\textbf{MetaWorld (Hard)}} & \multicolumn{3}{c}{\textbf{MetaWorld (Very Hard)}}   \\
\textbf{Method}   & Push Wall & Soccer                   & \multicolumn{1}{c|}{Sweep} & Pick Out of Hole   & Push             & Stick Pull  & Stick Push  & Pick Place Wall \\ 
\midrule
CILP      & 6         & 0     & \multicolumn{1}{c|}{6}    & 0    & 10    & \cellcolor[HTML]{ffe5cc}66    & \cellcolor[HTML]{ffcccb}100    & 0               \\
DINOv2    & 2         & 24   & \multicolumn{1}{c|}{0}     & 0     & 10   & \cellcolor[HTML]{ffe5cc}66    & \cellcolor[HTML]{ffcccb}100         & 0               \\
PointNet  & 40        & 24   & \multicolumn{1}{c|}{52}    & 10    & 36   & 30          & \cellcolor[HTML]{ffe5cc}88          & 48              \\
PointMAE  & 66        & 22   & \multicolumn{1}{c|}{88}    & 16                 & 68     & 58    & 76       & 76              \\
PointDif  & 38        & 26   & \multicolumn{1}{c|}{\cellcolor[HTML]{ffe5cc}92}    & 20     & 62     & 52          & 76          & 48              \\
Dynamo    & 26        & 26   & \multicolumn{1}{c|}{12}    & 16    & 12    & 16     & 72          & 14              \\
Dynamo-3D & 58        & 24   & \multicolumn{1}{c|}{88}    & 4   & \cellcolor[HTML]{ffcccb}78     & 60    & \cellcolor[HTML]{ffcccb}100    & \cellcolor[HTML]{ffe5cc}80              \\
FVP       & 36        & 34   & \multicolumn{1}{c|}{44}    & \cellcolor[HTML]{ffe5cc}26           & 42               & 24          & 84          & 78              \\
DP3       & \cellcolor[HTML]{ffe5cc}78        & \cellcolor[HTML]{ffcccb}38                       & \multicolumn{1}{c|}{\cellcolor[HTML]{ffe5cc}92}    & 24                 & \cellcolor[HTML]{ffe5cc}74               & 58          & \cellcolor[HTML]{ffe5cc}88          & \cellcolor[HTML]{ffcccb}94              \\
\textbf{AFRO (Ours)}  & \cellcolor[HTML]{ffcccb}80        & \cellcolor[HTML]{ffe5cc}36           & \multicolumn{1}{c|}{\cellcolor[HTML]{ffcccb}98}    & \cellcolor[HTML]{ffcccb}32                 & \cellcolor[HTML]{ffcccb}78               & \cellcolor[HTML]{ffcccb}78          & \cellcolor[HTML]{ffcccb}100         & \cellcolor[HTML]{ffcccb}94              \\ 
\bottomrule[2pt]
\end{tabular}}
\label{tab: complete metaworld results}
\end{table*}

\section{Further Details on RH20T}
\label{appendix:rh20t}
RH20T\cite{rh20t} (Robot--Human demonstration in 20TB) is a large-scale, real-world robotic manipulation dataset designed to support one-shot imitation learning and general skill learning across diverse, contact-rich tasks. The dataset contains over $110{,}000$ robot manipulation sequences paired with an equal number of human demonstration videos, resulting in more than $50$ million image frames in total and about $20$~TB of data. It demonstrates over 140 different tasks, including a wide variety of operational tasks.  Each robot sequence is recorded in the real world and includes synchronized multi-modal information such as RGB and depth images, force/torque measurements, audio, and low-level robot state and action signals, together with a corresponding human demonstration video and a language description for the same skill. We generate a point cloud for each frame by back-projecting the depth map into 3D space using the corresponding camera intrinsics. We then apply farthest point sampling (FPS) to downsample the raw point cloud to 1024 points. Empirically, we find that 1024 points provide sufficient geometric coverage to capture the object contours while at the same time substantially reducing the computational cost of the PointTransformer encoder.

\section{Limitations and Future Work}
Although AFRO achieves strong performance across diverse simulated and real-world manipulation tasks, several limitations remain.:contentReference[oaicite:0]{index=0}  
First, our current pre-training regime is moderate in scale compared with truly large vision foundation models: we use task-specific simulation data and a subset of RH20T rather than hundreds of millions of frames.  
Second, the objective is predominantly dynamics-driven.  
While this helps encode transition structure, the learned features still lack the rich semantic coverage that large 2D or 3D visual models obtain from web-scale data.  
Third, our framework is instantiated with single-view point clouds from a fixed depth camera; it does not yet exploit the multi-view nature of real 3D environments, such as fusing observations from multiple cameras over time.  
Fourth, the latent action is an abstract variable optimized only for predictive performance; it does not have explicit physical meaning (e.g., end-effector displacement, velocity, or contact force), which limits its interpretability and potential reuse.  
Finally, AFRO relies on VICReg-style variance regularization to prevent collapse in latent space; as shown in our ablations, removing this constraint severely harms performance, indicating that training stability is still tied to a relatively strong handcrafted objective.

These observations suggest several directions for future work.  
(1) \textbf{Scaling and semantic grounding.} We plan to combine AFRO with strong semantic backbones such as VGGT~\cite{wang2025vggt} or DINOv2 style models, either by distilling their features into the 3D encoder or by joint multi-task pre-training.  
This would aim to obtain representations that are simultaneously dynamics-aware and semantically rich.  
(2) \textbf{Larger-scale pre-training.} A natural extension is to train AFRO on substantially larger and more diverse 3D robot datasets, including multi-embodiment, multi-task, and web-scale synthetic data, to approach the scale of general-purpose visual foundation models for robotics.  
(3) \textbf{Multi-view 3D dynamics.} Incorporating multi-view inputs and explicitly modeling cross-view consistency of dynamics (e.g., by fusing trajectories from several cameras) could make the representation more robust to occlusion and viewpoint changes and better suited to mobile manipulation.  
(4) \textbf{Physically grounded latent actions.} We are interested in tying latent actions to physically interpretable quantities, for example by supervising them with estimated optical flow, 3D scene flow, or end-effector trajectories, so that the latent space captures meaningful motion primitives that can be reused across tasks and robots.  
(5) \textbf{Collapse-free objectives beyond VICReg.} Finally, we aim to explore alternative self-supervised objectives—such as JEPA-style prediction, information-theoretic regularization, or architectural constraints—that maintain feature diversity without relying on explicit variance penalties, potentially simplifying training and further improving robustness of the learned 3D representations.

\end{document}